\documentclass[graybox]{svmult}

\usepackage{type1cm}        
\usepackage{makeidx}         
\usepackage{graphicx}        
\usepackage{multicol}        
\usepackage[bottom]{footmisc}
\usepackage{breakcites}
\usepackage{url}
\usepackage{newtxtext}
\usepackage{newtxmath}       
\usepackage[normalem]{ulem}

\makeindex             
\usepackage{array}
\usepackage{subcaption}
\usepackage{comment}
\usepackage{multirow}
\def\mystrut(#1,#2){ \vrule height #1 depth #2 width 0pt}
\newcolumntype{C}[1]{%
   >{\mystrut(1ex,1.5ex)\centering}%
   m{#1}%
   <{}}
\renewcommand{\arraystretch}{1}

\begin{document}

\title*{Influence of Color Spaces for Deep Learning Image Colorization}

\author{Coloma Ballester, Aurélie Bugeau, Hernan Carrillo, Michaël Clément, Rémi Giraud, Lara Raad, Patricia Vitoria}
\authorrunning{C. Ballester, A. Bugeau, H. Carrillo, M. Clément, R. Giraud, L. Raad, P. Vitoria}

\institute{Coloma Ballester (\email{coloma.ballester@upf.edu}) \at Universitat Pompeu Fabra, Barcelona, Spain.
\and Aurélie Bugeau (\email{aurelie.bugeau@labri.fr}) \at Univ. Bordeaux, LaBRI, CNRS UMR 5800, France.
\and Hernan Carrillo (\email{hernan.carrillo-lindado@u-bordeaux.fr}) \at Univ. Bordeaux, LaBRI, CNRS UMR 5800, France.
\and Michaël Clément (\email{michael.clement@labri.fr}) \at Univ. Bordeaux, Bordeaux INP, LaBRI, CNRS UMR 5800, France.
\and Rémi Giraud (\email{remi.giraud@ims-bordeaux.fr}) \at Univ. Bordeaux, Bordeaux INP, IMS, CNRS UMR 5218, France.
\and  Lara Raad (\email{lara.raadcisa@esiee.fr}) \at Université Paris-Est, LIGM (UMR 8049), CNRS, ENPC, ESIEE Paris, UPEM, France.
\and Patricia Vitoria (\email{patricia.vitoria@upf.edu}) \at Universitat Pompeu Fabra, Barcelona, Spain.}

\maketitle

\vspace{-2cm}

\abstract{Colorization is a process that converts a grayscale image into a color one that looks as natural as possible. Over the years this task has received a lot of attention. Existing colorization methods rely on different color spaces: RGB, YUV, Lab, etc. In this chapter, we aim to study their influence on the results obtained by training a deep neural network, to answer the question: "Is it crucial to correctly choose the right color space in deep-learning based colorization?". First, we briefly summarize the literature and, in particular, deep learning-based methods. We then compare the results obtained with the same deep neural network architecture with RGB, YUV and Lab color spaces. Qualitative and quantitative analysis do not conclude similarly on which color space is better. We then show the importance of carefully designing the architecture and evaluation protocols depending on the types of images that are being processed and their specificities: strong/small contours, few/many objects, recent/archive images.}

\section{Introduction}
\label{sec:intro} 

Image colorization consists in recovering a color image from a grayscale one.
This process attracts a lot of attention in the image-editing community in order to restore or colorize old grayscale movies or pictures.
While turning a color image into a grayscale one is only a matter of standard, the reverse operation is a strongly ill-posed problem as no information on which color has to be added is known.
Therefore priors must be considered.
In the literature, there exist three kinds of priors leading to different types of colorization methods.
In the first category, initiated by~\cite{Levin2004}, the user manually adds  initial colors through scribbles to the grayscale image.
The colorization process is then performed by propagating the input color data to the whole image.
The second category, called automatic or patch-based colorization, initiated by~\cite{Welsh2002}, consists in transferring color from one (or many) initial color image considered as example.
The last category, which attracts most research nowadays, concerns deep learning approaches.
The necessary color prior here is learned from large datasets.

Generally, in colorization methods, the initial grayscale image is considered as the luminance channel which is not modified during the colorization.
The objective is then to reconstruct the two chrominance channels, before turning back to the RGB color space.
Different luminance-chrominance spaces exist and have been used for image colorization.
One common problem with all image colorization methods that aim at reconstructing the chrominances of the target image is that the recovered chrominances combined with the input luminance may not fall into the RGB cube when converting back to the RGB color space.
Therefore, some works have decided to work directly in RGB to cope with this limitation by constraining the luminance channel~\cite{Pierre2014b}.

The objective of this chapter is to analyze the influence of color spaces on the results of automatic deep learning methods for image colorization.
This chapter comes together with another chapter of this handbook.
This other chapter, called \textit{Analysis of Different Losses for Deep Learning Image Colorization}, focuses on the influence of losses.
We refer the reader to it for a review of the traditionally used different losses and evaluation metrics.
Here, after reviewing existing works in image colorization and, in particular, works based on deep learning, we will focus on the influence of color spaces.
Based on our analysis of the literature, a baseline architecture is defined and later used in all comparisons.
Additionally, again based on the literature review, we set a uniform training procedure to ensure fair comparisons.
Experiments encompass qualitative and quantitative analysis.

The chapter is organized as follows.
Section~\ref{sec:sota} first recalls some basics on color spaces, then provides a detailed survey of the literature on colorization methods and finally lists the datasets traditionally used.
Next, in Section~\ref{sec:framework}, we present the chosen architecture and in Section~\ref{sec:learning} the learning strategy.
Section~\ref{sec:colorinfluence} presents the results of the different experiments.
A discussion on the generalization of this work to archive images is later provided in Section~\ref{sec:archive} before a conclusion is drawn.

\section{Related Work}
\label{sec:sota}

\subsection{On Color Spaces}

This section presents the different color spaces that have been used for colorization in the literature.
For more information about color theory and color constancy (i.e., the underlying ability of human vision to perceive colors very robustly with respect to changes of illumination) see, for instance \cite{ebner2007color,fairchild2013color}.

Color images are traditionally saved in the RGB color space.
A grayscale image contains only one channel that encodes the luminosity (perceived brightness of that object by a human observer) or the luminance (absolute amount of light emitted by an object per unit area).
A way to model this luminance $Y$ which is close to the human perception of luminance is:
\begin{equation}\label{def_lum}
Y = 0.299 R + 0.587 G + 0.114 B,
\end{equation}
where $R, G$ and $B$ are, respectively, the amount of light emitted by an object per unit area in the low, medium and high frequency bands that are visible by a human eye.
Colorization aims to retrieve color information from a grayscale image.
To do so, and to easily constrain the luminance channel, most methods propose to work in a luminance-chrominance space.
The problem becomes the retrieval of two chrominance channels given the luminance $Y$.
There exist several luminance-chrominance spaces.
Two of them are mostly used for colorization.
The first one, YUV, historically used for a specific analog encoding of color information in television systems, is the result of the linear transformation:
$$
\left(\begin{array}{c}
Y \\
U \\
V
\end{array} \right)
=
\left( \begin{array}{ccc}
0.299 & 0.587 & 0.114 \\
-0.14713 & -0.28886 & 0.436 \\
 0.615 & -0.51498 & -0.10001
\end{array} \right)
\left(\begin{array}{c}
R \\
G \\
B
\end{array} \right). $$
The reverse conversion from YUV and RGB is simply obtained by inverting the matrix.
The other linear space that has been used for colorization is YCbCr.

The CIELAB color space, also referred to as Lab or La$^\star$b$^\star$, defined by the International Commission on Illumination (CIE) in 1976, is also frequently used for colorization.
It has been designed such that the distances between colors in this space correspond to the perceptual distances of colors for a human observer.
The three channels become uncorrelated. The transformation from RGB to Lab (and the reverse) is non linear.
First, it is necessary to convert the RGB values to the CIEXYZ color space:
$$
\left(\begin{array}{c}
X \\
Y \\
Z
\end{array} \right)
=
\left( \begin{array}{ccc}
2.769& 1.7518 & 0.13 \\
1 & 4.5907 & 0.0601 \\
 0 & 0.0565 & 5.5943
\end{array} \right)
\left(\begin{array}{c}
R \\
G \\
B
\end{array} \right). $$
Then, the transformation to Lab is given by
$$\begin{array}{ll}
  L &= 116 f(Y/Y_n) - 16,\\
  a &= 500 \left[f(X/X_n) - f(Y/Y_n)\right],\\
  b &= 200 \left[f(Y/Y_n) - f(Z/Z_n)\right],
\end{array}
$$
with $$f(t) = \begin{cases} 
  t^{1/3} & \mbox{if } t > (\frac{6}{29})^3, \\
  \frac13 \left( \frac{29}{6} \right)^2 t + \frac{4}{29} & \mbox{otherwise},
\end{cases}$$ 
where $X_n$, $Y_n$ and $Z_n$ describe a specified white achromatic reference illuminant. Obviously, the reverse operation from Lab to RGB is also non linear. \\

Despite RGB or luminance-chrominance color spaces, few methods relying on hue-based spaces have been proposed for colorization. For instance, \cite{larsson2016learning} rely on a hue-chroma-luminance space.\\

Table~\ref{tab:colors} lists the color spaces used in deep learning colorization methods described in the next subsection.
It distinctly appears that the Lab color space is the most widely used.
We will further discuss this choice in Section~\ref{sec:colorinfluence}.

\begin{table}[h!]
\centering
{\footnotesize
\hspace{-0.5cm}
\begin{tabular}{|c|c|c|c|c|c|c|c|c|c|c|c|c|c|c|c|c|c|c|c|c|}
\hline
&&&\multicolumn{5}{c|}{Using}&\multicolumn{3}{c|}{Histogram}&\multicolumn{2}{c|}{User}&\multicolumn{4}{c|}{Diverse}&\multicolumn{3}{c|}{Object}&\multicolumn{1}{c|}{Survey}\tabularnewline
&&&\multicolumn{5}{c|}{GANs}&\multicolumn{3}{c|}{prediction}&\multicolumn{2}{c|}{guided}&\multicolumn{4}{c|}{}
&\multicolumn{3}{c|}{aware}
&\multicolumn{1}{c|}{}\tabularnewline
\hline
&\rotatebox[origin=c]{90}{\cite{Cheng2015}}
&\rotatebox[origin=c]{90}{\cite{Iizuka2016}}
&\rotatebox[origin=c]{90}{\cite{vitoria2020chromagan}}
&\rotatebox[origin=c]{90}{\cite{nazeri2018image}}
&\rotatebox[origin=c]{90}{\cite{cao2017unsupervised}}
&\rotatebox[origin=c]{90}{\cite{yoo2019coloring}}
&\rotatebox[origin=c]{90}{\cite{antic2019deoldify}}
&\rotatebox[origin=c]{90}{\cite{larsson2016learning}}
&\rotatebox[origin=c]{90}{\cite{Zhang2016}}
&\rotatebox[origin=c]{90}{\cite{mouzon2019joint}}
&\rotatebox[origin=c]{90}{\cite{zhang2017real}}
&\rotatebox[origin=c]{90}{\cite{he2018deep}}
&\rotatebox[origin=c]{90}{\cite{Deshpande2017}}
&\rotatebox[origin=c]{90}{\cite{guadarrama2017pixcolor}}
&\rotatebox[origin=c]{90}{\cite{royer2017probabilistic}}
&\rotatebox[origin=c]{90}{\cite{kumar2021colorization}}
&\rotatebox[origin=c]{90}{\cite{su2020instance}}
&\rotatebox[origin=c]{90}{\cite{pucci2021collaborative}}
&\rotatebox[origin=c]{90}{\cite{kong2021adversarial}}
&\rotatebox[origin=c]{90}{\cite{Gu_2019_CVPR_Workshops} (winner)}
\tabularnewline
\hline
RGB&&&&&$\bullet$&$\bullet$&$\bullet$&&&&&&&&&$\bullet$&&&&$\bullet$\tabularnewline
\hline
YUV&$\bullet$&&&&$\bullet$&&$\bullet$&&&&&&&&&&&&&\tabularnewline
\hline
YCbCr&&&&&&&&&&&&&&$\bullet$&&&&&&\tabularnewline
\hline
Lab&&$\bullet$&$\bullet$&$\bullet$&&$\bullet$&&$\bullet$&$\bullet$&$\bullet$&$\bullet$&$\bullet$&$\bullet$&&$\bullet$&&$\bullet$&$\bullet$&$\bullet$&\tabularnewline
\hline
hue/chroma&&&&&&&&$\bullet$&&&&&&&&&&&&\tabularnewline
\hline
Comparison&&$\bullet$&&&$\bullet$&&&$\bullet$&&&&&&&&&&&&\tabularnewline
\hline

\end{tabular}}
\caption{Color spaces used in deep learning methods for image colorization.}\label{tab:colors}
\end{table}

In general terms, as can be seen in Table~\ref{tab:colors}, most methods work in a luminance-chrominance space and the cost functions to optimize are in general defined in the same space.
Hence, converting from and to RGB to one of these luminance/chrominance spaces is not involved in the backpropagation step.
Once the training is performed, at inference time the chrominance values given by the network together with the luminance component are converted back to the RGB color space.
As mentioned earlier, this operation tends to perform an abrupt value clipping to fit in the RGB cube hence modifying both the original luminance values and the predicted chrominance values.
Two libraries are most commonly used for the conversion step: the color module of scikit-image (\cite{Zhang2016,larsson2016learning,zhang2017real,royer2017probabilistic}) and the color space conversions functions of OpenCV (\cite{Iizuka2016,vitoria2020chromagan}).

\subsection{Review of Colorization Methods}
\label{sec:review}

This section presents an overview of the colorization methods in the three categories: scribble-based, exemplar-based and deep learning.
For a more detailed review with the same classification, we refer the reader to the recent review~\cite{li2020review}.
Another survey focused on deep learning approaches proposes a taxonomy to separate these methods into seven categories \cite{anwar2020image}.
The authors of this review have redrawn all networks architectures thus allowing to easily compare architecture specificity.
Comparisons of methods are made on a new Natural-Color Dataset made of objects with white background.

The NTIRE challenge is a competition for different computer vision tasks related to image enhancement and restoration.
One of the tasks in 2019 was image colorization~\cite{Gu_2019_CVPR_Workshops}, with two tracks: colorization without or with guidance given by a second input that provides several color guiding points.

\subsubsection{Scribble-based Image Colorization}

\begin{figure}[h!]
\centering
\begin{tabular}{cc}
\includegraphics[ width=0.45\textwidth]{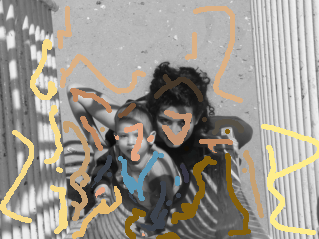} &
\includegraphics[ width=0.45\textwidth]{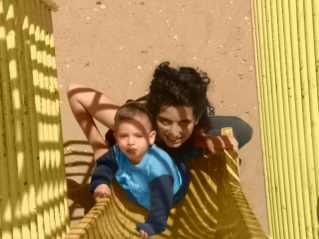}
\end{tabular}
\caption{Example of scribble-based image colorization taken from ~\cite{Levin2004}. The user draws color that are successively diffused to neighbor pixels according under some constraints that depend on the different methods. }
\label{fig:scrib}
\end{figure}

The first category of colorization methods relies on color priors coming from scribbles drawn by the user (see Figure~\ref{fig:scrib}).
These colors are propagated to all pixels by diffusion schemes.

The first manual colorization method based on scribbles was proposed by \cite{Levin2004}.
It solves an optimization problem to diffuse the chrominances of scribbles with the assumption that chrominances should have small variations where the luminance has small variations. 
To reduce the number of needed scribbles, \cite{luan2007natural} first use scribbles to segment the image before diffusing the colors.
\cite{Yatziv2006} propose a simple yet fast method by using geodesic distances to blend the chrominances given by the scribbles.
In \cite{huang2005adaptive}, edge information is extracted to reduce color bleeding.
\cite{Heu2009} use pixel priorities to ensure that important areas end up with the right colors.
Other propagation schemes include probabilistic distance transform \cite{Lagodzinski2008}, discriminative textural features \cite{Kawulok2012}, structure tensors~\cite{Drew2011}, non local graph regularization \cite{Lezoray2008}, matrix completion~\cite{wang2012colorization, yao2015colorization} or rank minimization~\cite{ling2015image}.
As often described in the literature, with these manual approaches, the contours are not well preserved.
To cope with this issue, in \cite{Ding2012}, scribbles are automatically generated after segmenting the image and the user only needs to provide one color per scribble.
However, all manual methods suffer from the following drawback: if the target represents a complex scene, the user interaction becomes very important.
On the other hand, these approaches propose a global optimization over the image, thus leading to spatial consistency in the result.

\subsubsection{Exemplar-based Image Colorization}

\begin{figure}[h!]
\centering
\includegraphics[width=\textwidth]{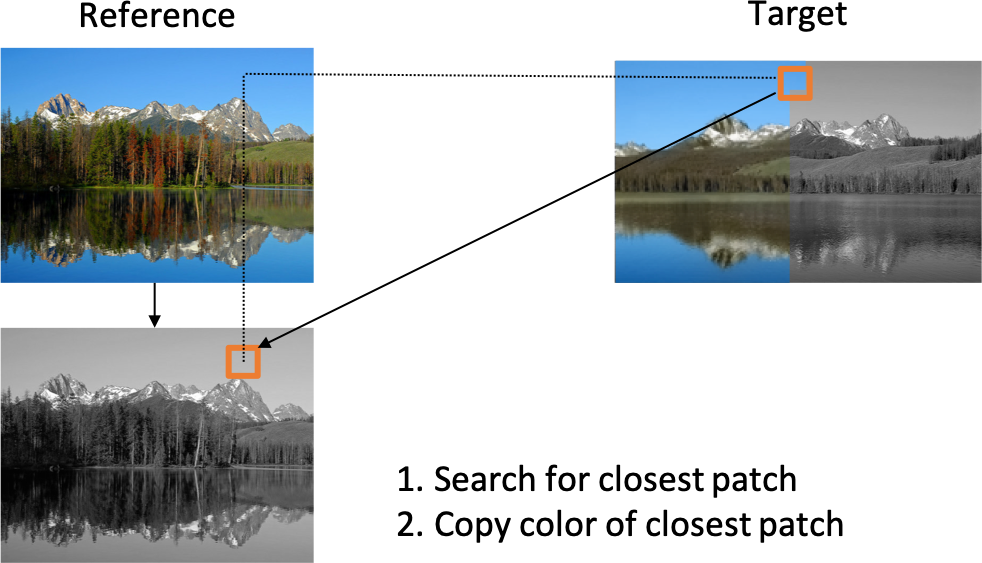}
\caption{Principle of exemplar-based image colorization. Methods in this category have proposed different similar patch search strategies, and techniques to add spatial consistency when copying patch colors.}
\label{fig:ex}
\end{figure}

The second category of colorization methods concerns exemplar-based methods which rely on a color reference image as prior.
The first exemplar-based colorization method was proposed by \cite{Welsh2002}.
It makes the assumption that pixels with similar intensities or similar neighborhood should have similar colors.
It extends the texture synthesis approach by \cite{Efros1999}: the final color of one pixel is copied from the most similar pixel in a reference input color image.
The similarity between pixels relies on patch-based metrics (see Figure~\ref{fig:ex}).
This approach has given rise to many extension in the literature~\cite{DiBlasi2003, liu2012automatic}.
In particular, many works have focused on choosing or designing appropriate features for matching pixels \cite{Chia2011,Gupta2012, Bugeau2012,Cheng2015, Arbelot2016, Arbelot2017}.

To overcome the spatial consistency and coupling problems in automatic methods, several works rely on image segmentation.
For instance, \cite{Irony2005} propose to determine the best matches between the target pixels and regions in a pre-segmented source image.
With these correspondences, micro-scribbles from the source are initialized on the target image and colors are propagated as in~\cite{Levin2004}.
\cite{tai2005local} build a probabilistic segmentation of both images where one pixel can belong to many regions.
They use it to transfer color between any two regions having similar statistics with an Expectation-Maximization scheme.
\cite{Gupta2012} extract different features from the superpixels~\cite{Ren2003} of the target image and match them with the source ones.
The final colors are computed by imposing spatial consistency as in~\cite{Levin2004}.
\cite{li2017example} extract low and high-level features on superpixels of the reference to form a dictionary then used as a dictionary-based sparse reconstruction problem.
Sparse representation was previously used for colorization in~\cite{pang2013image} where images are segmented from scribbles.
These approaches incorporate local consistency into automatic methods via segmentation.
In~\cite{Charpiat2008}, spatial consistency is solved with graph cuts after estimating for each pixel the conditional probability of colors.
In~\cite{Bugeau2014, Pierre2014b} each pixel can only take its chrominance (or RGB color) among a reduced set of possible candidates chosen from the reference image.
The final color is chosen using a variational formulation.
In the same trend, \cite{fang2019superpixel} proposes a superpixel based variational model.
In~\cite{li2017exampleb}, the distribution of intensity deviation for uniform and non-uniform regions is learned and used in a Markov Random Field (MRF) model for improved consistency.
Finally,~\cite{li2019automatic} propose cross-scale local texture matching, which are then fused using global graph-cut optimisation.

A major problem of this family of methods is the high dependency on the reference image.
\cite{Chia2011} therefore propose to rely on several reference images obtained from an internet search based on semantic information.

\subsubsection{Deep Learning Methods for Image Colorization}
Since 2012, deep learning approaches, and in particular Convolutional Neural Networks (CNNs), have become very popular in the community of computer vision and computer graphics.

The first deep learning-based colorization methods were proposed in~\cite{Cheng2015, Deshpande2015}.
In~\cite{Cheng2015}, a fully automated system extracts handcrafted low and high features and feeds them as input to a three-layer fully connected neural network trained with a L2 loss.
The network predicts the U and V channels of the YUV luminance-chrominance space.
The authors also add an optional clustering stage where the images are divided in different types of scenes, according to the previously extracted semantic features.
Then, a different neural network is trained for each of the clusters. \\

\begin{figure}[h!]
\centering
\includegraphics[ width=\textwidth]{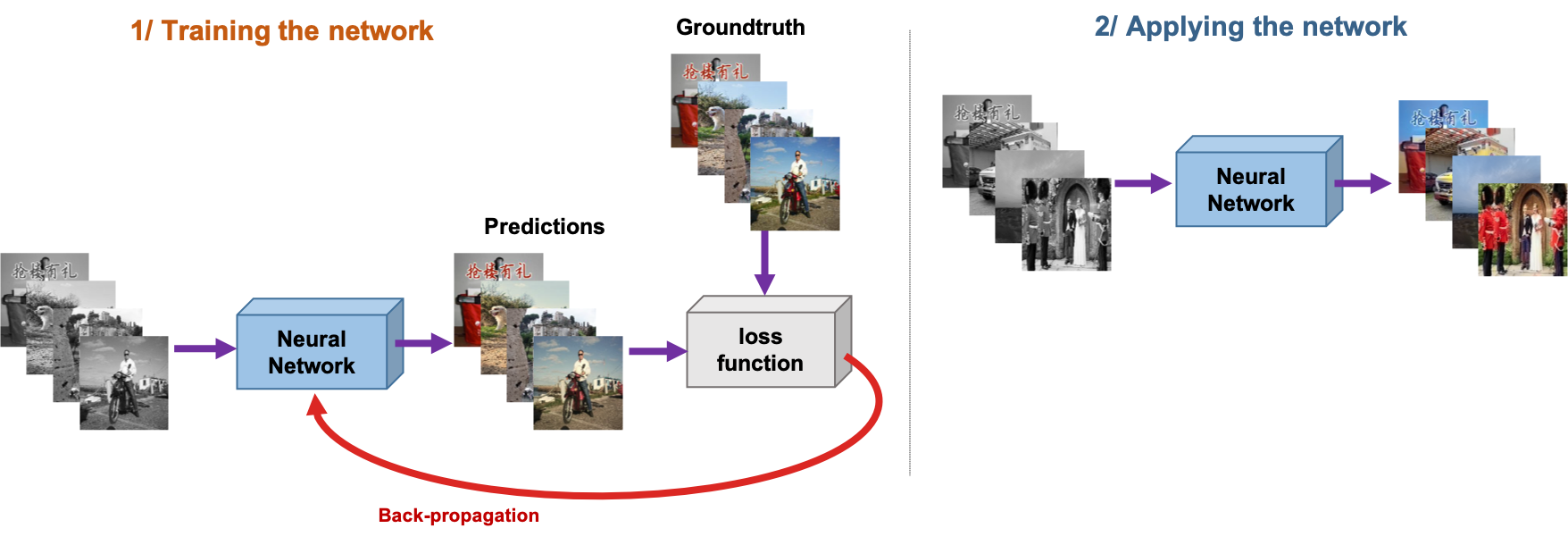}
\caption{Principle of basic end-to-end colorization networks.}
\label{fig:dl}
\end{figure}

\textbf{End-to-end approaches: } Later on, papers focused more on \textit{end-to-end approaches} (see Figure~\ref{fig:dl}).

For instance, the paper that won both tracks of the \cite{Gu_2019_CVPR_Workshops} NTIRE 2019 Challenge on Image Colorization was the end-to-end method proposed by IPCV\_IIMT.
It implements an encoder-decoder structure that resembles to a {U-Net} with the encoder built using deep dense-residual blocks.
\cite{wan2020automated} proposes to combine neural networks with color propagation.
It first trains a neural network in order to colorize interest points of extracted superpixels.
Then those colors are propagated by optimizing an objective function.
In an older work, \cite{Iizuka2016} presented an end-to-end colorization framework based on CNNs to infer the \textit{ab} channels of the CIE Lab color space.
This work is built on the basis that a classification of the images can help to provide global priors that will improve the colorization performance.
The network extracts global and local features and is jointly trained for classification and colorization in a labeled dataset. \\

\textbf{Using GANs:} Still being end-to-end, other methods use \textit{generative adversarial networks} (GANs)~\cite{goodfellow2014generative}.
\cite{isola2017image} propose the so-called image-to-image method pix2pix.
It maps an input image to an output image using a U-Net generator and a patch GANs discriminator.
The method is used in many applications including colorization.
This method was extended in~\cite{nazeri2018image} using Deep Convolutional GANs (DCGAN)~\cite{dcgan}.
In \cite{cao2017unsupervised}, a fully convolutional generator with a conditional GANs is considered.
This architecture does not use downsampling to avoid extracting global features which are not suitable to recover accurate boundaries.
To avoid noise attenuation and make the colorization results more diversified, they concatenate a noise channel onto the first half of the generator layers.
GANs have also been used in chromaGAN~\cite{vitoria2020chromagan} which extends \cite{Iizuka2016} by proposing to learn the semantic image distribution without any need of a labeled dataset.
This method combines three losses: a color error loss by computing MSE on \textit{ab} channels, a class distribution loss by computing the Kullback-Leibler divergence on VGG-16 class distribution vectors, and an adversarial Wasserstein GAN (WGAN) loss \cite{arjovsky2017wasserstein}.
To prevent the need for training on a huge amount of data,~\cite{yoo2019coloring} introduce MemoPainter, a few-shot colorization framework.
MemoPainter is able to colorize an image with limited data by using an external memory network in addition to a colorization network.
The memory network learns to retrieve a color feature that best matches the ground-truth color feature of the query image, while the generator-discriminator colorization network learns to effectively inject the color feature to the target grayscale image.

DeOldify~\cite{antic2019deoldify} is another end-to-end image and video colorization method mapping the missing chrominance values to the grayscale input image.
A ResNet (ResNet101 or ResNet34) is used as the backbone of the generator of a U-Net architecture trained as follows : the generator is first trained with the perceptual loss~\cite{johnson2016perceptual}, followed by training the critic as a binary classsifier distinguishing between real images and those generated by the generator and finally the generator and critic are trained together in an adversarial manner on $1-3\%$ of the ImageNet~\cite{deng2009imagenet} data.
The latter is the so-called NoGAN strategy which is enough to add color realism to the results and which also allows to avoid flickering across video frames while the colorization is applied individually frame per frame.

\textbf{Predicting distributions instead of images:} Regression does not handle multimodal color distributions well~\cite{larsson2016learning}.
\cite{larsson2016learning, Zhang2016} address this issue by \textit{predicting distributions} over a set of bins, as it was initially done in the exemplar-based method~\cite{Charpiat2008}.
They therefore rely on a discretization of color spaces.
In \cite{larsson2016learning}, the color space is binned with evenly spaced Gaussian quantiles.
Experiments are run for hue/chroma and Lab colors spaces with either separated or joint distributions.
Inference of the colored image from the distribution uses expectation (sum over the color bin centroids weighted by the histogram).
In~\cite{Zhang2016}, the \textit{ab} output space is quantized into bins with grid size 10 and the 313 values which are in gamut are kept.
The inference is the annealed-mean of the distribution.
In~\cite{mouzon2019joint, pierre2020recent}, the resulting distributions from~\cite{Zhang2016} are later used in a variational approach~\cite{Pierre2015d}.

\textbf{Considering user priors:} Few methods give the possibility to add user inputs as additional priors.
The architecture in~\cite{zhang2017real} learns to propagate color hints by fusing low-level cues and high-level semantic information.
\cite{he2018deep} uses a reference color image to guide the output of their deep exemplar-based colorization method.

\newcommand\tab[1][1cm]{\hspace*{#1}}

{\scriptsize
\begin{table}[h!]
\renewcommand{\arraystretch}{1.1}
\centering
{\scriptsize
\begin{tabular}{@{\hspace{0mm}}|C{2.2cm} | C{2.2cm} | C{2.25cm} | C{2.25cm} | C{2.25cm} |@{\hspace{0mm}}}
\hline
& {\footnotesize Additional inputs} & {\footnotesize Network} & {\footnotesize Network's output} & {\footnotesize Post-processing}\tabularnewline
\hline
  \cite{Cheng2015} & hand-crafted features & 3 layers FC & UV & joint bilateral filtering\tabularnewline
\hline
  \cite{Iizuka2016} & $-$ & CNNs (local/global) & \textit{ab} & upsampling \tabularnewline
\hline
  \cite{wan2020automated} & superpixels' hand-crafted features & FC net & interest points' color & propagation and refinement \tabularnewline
\hline

\multicolumn{5}{|c|}{\textbf{Using GANs}}\tabularnewline
\hline
  \cite{vitoria2020chromagan} & $-$ & CNNs (local/global) + PatchGAN & \textit{ab} & upsampling \tabularnewline
\hline
  \cite{nazeri2018image} & $-$ & U-Net~\cite{isola2017image} + DCGAN & Lab & $-$ \tabularnewline
\hline
  \cite{cao2017unsupervised} & $-$ & FCONV generator with multi-layer noise + PatchGAN & UV/RGB (diverse) & $-$ \tabularnewline
\hline
  \cite{yoo2019coloring} & color thief features & colorisation U-Net + memory nets noise & $-$ & $-$ \tabularnewline
\hline
  \cite{antic2019deoldify} & $-$ & U-Net + self-attention + GAN & RGB & YUV conversion + cat(original Y/UV) + RGB conversion \tabularnewline

\hline
\multicolumn{5}{|c|}{\textbf{Histograms Prediction}}\tabularnewline
\hline
  \cite{larsson2016learning} & $-$ & VGG-16 + FC layers & distributions & expectation \tabularnewline
\hline
  \cite{Zhang2016} & $-$ & VGG-styled net & distributions & annealed mean \tabularnewline
\hline
  \cite{mouzon2019joint} & $-$ & \cite{Zhang2016} & distributions & variational model \tabularnewline
\hline
\multicolumn{5}{|c|}{\textbf{User guided}}\tabularnewline
\hline
  \cite{zhang2017real} & user point, global histograms and average saturation & U-Net & distributions + \textit{ab} & $-$ \tabularnewline
\hline
  \cite{he2018deep}& color reference & similarity sub-net + U-Net (gray VGG-19) & bidirectional similarity maps + \textit{ab} & $-$ \tabularnewline
\hline
\multicolumn{5}{|c|}{\textbf{Diverse colorization and autoregressive models}}\tabularnewline
\hline
  \cite{Deshpande2017} & $-$ & cVAE + MDN & diverse colorization & $-$ \tabularnewline
\hline
  \cite{guadarrama2017pixcolor} & $-$ & PixelCNN + CNN & diverse colorization & $-$ \tabularnewline
\hline
  \cite{royer2017probabilistic} & $-$ & CNN + PixelCNN++ & diverse colorization & $-$ \tabularnewline
\hline
  \cite{kumar2021colorization} & $-$ & axial transformer + color/spatial upsamplers (self-attention blocks) & diverse colorization & $-$ \tabularnewline
\hline
\multicolumn{5}{|c|}{\textbf{Object aware}}\tabularnewline
\hline
  \cite{su2020instance}& object bounding boxes & U-Net (global/instance) + CNN (fusion) & \textit{ab} & $-$ \tabularnewline
\hline
  \cite{pucci2021collaborative} & $-$ & CNN + capsule net & \textit{ab} & $-$ \tabularnewline
\hline
  \cite{kong2021adversarial} & $-$ & U-Net + PatchGAN & \textit{ab} + semantic segmentation & $-$ \tabularnewline
\hline
\multicolumn{5}{|c|}{\textbf{Survey}}\tabularnewline
\hline
  \cite{Gu_2019_CVPR_Workshops} & $-$ & U-Net & RGB & $-$ \tabularnewline
\hline
\end{tabular}}
\caption{Short description of deep networks for image colorization, their input other than grayscale image, output. Here FCONV stands for fully convolutional, FC for fully connected and U-Net for a U-Net-like network and not the vanilla U-Net.}
\label{tab:nets}
\end{table}}

\textbf{Generating diverse image colorizations:} Some methods have been designed to generate diverse colorizations as there is not one unique solution to the colorization problem.
\cite{Deshpande2017} relies on a variational auto-encoder (VAE) to learn a low dimensional embedding of color spaces.
The mapping from a grayscale input image to color distribution of the latent space is done by learning a mixture density network (MDN).
At test time, it is possible to sample the conditional model and use the VAE decoder to generate diverse color images.
In their PixColor model, \cite{guadarrama2017pixcolor} first train a conditional PixelCNN \cite{oord2016conditional} to generate multiple latent low resolution color images, then train a second CNN to generate the final high resolution images.
Another method, called PIC, that uses PixelCNN++~\cite{salimans2017pixelcnn++} (an extension to the original PixelCNN), was proposed in \cite{royer2017probabilistic}.
A feed-forward CNN first maps grayscale image to an embedding that encodes color information.
This embedding is then fed to the autoregressive PixelCNN++ model which predicts a distribution of image chromacity.
The colTran model proposed by~\cite{kumar2021colorization}
is based on an axial transformer~\cite{ho2019axial} autoregressive model.
ColTran includes three networks, all relying on column/row self-attention blocks: the autoregressive model that estimates low resolution coarse colorization, a color upsampler and a spatial upsampler.

\textbf{Restoring and colorizing:} \cite{luo2020time} propose to specifically restore and colorize old black and white portrait photos in a unified framework.
It uses an additional high quality color reference image (the sibling) automatically generated by first training a network that projects images into the StyleGAN2~\cite{karras2020analyzing} latent space and then uses the pre-trained StyleGAN2 generator to create the sibling.
Fine details and colors are extracted from the sibling.
A latent code is then optimized through a three terms cost function and decoded by a StyleGAN2 generator yielding a high quality color version of the antique input.
The cost function is composed of a color term inspired by the style loss in~\cite{gatys2016image} between the features of the sibling and those of the generated high quality color image, a perceptual term~\cite{johnson2016perceptual} between a degraded version of the generative model's output and the antique input, and a contextual term between the VGG features of the sibling and those of the generated high quality color image.

\textbf{Decomposing the scene into objects:} Recently, some methods try to explicitely deal with the decomposition of the scene into objects in order to tackle one of the main drawbacks of most deep learning based colorization methods which is color bleeding across different objects.
\cite{su2020instance} proposes to colorize a grayscale image in an instance-aware fashion.
They train three separate networks: a first one that performs global colorization, a second one for instance colorization and a third one that fuses both colorization networks.
These networks are trained by minimizing the Hubber loss (also called Smooth L1 loss).
In general, after fusing both results the global colorization will be enhanced.
The instances per image are obtained by using a standard pre-trained object detection network, Mask R-CNN~\cite{he2017mask}.
\cite{pucci2021collaborative} propose to improve~\cite{Zhang2016} by using a network which is more aware of image instances, in the spirit of~\cite{su2020instance}, by combining convolutional and capsule networks.
They train from end-to-end a single network which first generates a per-pixel color distribution followed by a final convolutional layer that recovers the missing chrominance channels as opposed to~\cite{Zhang2016} that computes the annealed mean on the per-pixel color distribution network's output.
They train the network by minimizing the cross-entropy between per pixel color distributions and L2 loss on the chrominance channels.
\cite{kong2021adversarial} propose to colorize a grayscale image by training a multitask network for colorization and semantic segmentation in an adversarial manner.
They train a U-Net type network with a three term cost function: a color regression loss in terms of hue, saturation and lightness, the cross-entropy on the ground truth and generated semantic labels, and a GANs term.
The main objective of the proposal is to reduce color bleeding across edges.

Table~\ref{tab:nets} summarizes all these deep learning methods providing details on their particular inputs (other than the obvious grayscale image), their outputs, their architectures and pre- and post-processing steps.
This summary table is only provided for deep learning-based methods since we focus on deep learning-based strategies in the remaining of the chapter.

\subsection{Datasets used in Literature}

To train and test the deep-learning methods presented in previous subsection, different datasets have been used.
Table~\ref{table:datasets} summarizes the use of these datasets in colorization methods.
They contain from one thousand (DIV2K~\cite{agustsson2017ntire}) to million of images (ImageNet~\cite{deng2009imagenet}).
Image dimensions also vary a lot, from 32 $\times$ 32 in CIFAR-10~\cite{krizhevsky2009learning} to 2K resolution in DIV2K.

Other differences concern the content of the images itself.
Some datasets are very specific to a type of image: faces (LFW~\cite{LFWTech}), bedrooms (LSUN~\cite{yu2015lsun}).
Other present various scenes as Places~\cite{zhou2017places} with $205$ scene categories, COCO~\cite{lin2014microsoft} with $80$ object categories and $91$ stuff categories, and SUN~\cite{xiao2010sun} with $899$ scene categories.

\begin{table}[h!]
\centering
{\scriptsize
\hspace{-0.5cm}
\begin{tabular}{| c| c|c| c| c| c| c|c| c|c| C{4cm} |}
\hline
 &\rotatebox[origin=c]{90}{SUN} & \rotatebox[origin=c]{90}{ImageNet / ILSVRC-2015}& \rotatebox[origin=c]{90}{COCO} & \rotatebox[origin=c]{90}{CIFAR-10} & \rotatebox[origin=c]{90}{DIV2K} & \rotatebox[origin=c]{90}{Pascal VOC} & \rotatebox[origin=c]{90}{Places} & \rotatebox[origin=c]{90}{LSUN bedroom or church} 
 &\rotatebox[origin=c]{90}{testing on historic BW photo} &Remark / Other \tabularnewline
 
 \hline
\cite{Cheng2015} &$\bullet$& &&&& &&&&\tabularnewline
\hline
\cite{Iizuka2016} && &&&& &$\bullet$&&$\bullet$&\tabularnewline
\hline
\multicolumn{11}{|c|}{Using GANs}\tabularnewline
\hline
\cite{vitoria2020chromagan} && $\bullet$ &&&& &&&$\bullet$&\tabularnewline
\hline
\cite{nazeri2018image}& & & &$\bullet$ & &&$\bullet$&&&\tabularnewline
\hline
\cite{cao2017unsupervised}& & &&&& &&$\bullet$&& \tabularnewline
\hline
\cite{yoo2019coloring}& & &&&& &&&& Yumi, Monster, etc. \tabularnewline
\hline
\cite{antic2019deoldify} && $\bullet$ &&&& &&&&training on 1-3\% of ImageNet images\tabularnewline
\hline
\multicolumn{11}{|c|}{Histograms prediction}\tabularnewline
\hline
\cite{larsson2016learning} &$\bullet$& $\bullet$&&&& &&&&\tabularnewline
\hline
\cite{Zhang2016} && $\bullet$ &&&& $\bullet$&&&&training on 1.3M ImageNet images\tabularnewline
\hline
\multicolumn{11}{|c|}{User guided}\tabularnewline
\hline
\cite{zhang2017real}&&$\bullet$ &&&&&&&& \tabularnewline
\hline
\cite{he2018deep}&& $\bullet$ &&&&& &&&training on 700k ImageNet image/7 categories \tabularnewline
\hline
\multicolumn{11}{|c|}{Diverse}\tabularnewline
\hline
\cite{Deshpande2017}&&$\bullet$&&&&&&$\bullet$&&LFW \tabularnewline
\hline
\cite{guadarrama2017pixcolor}&&$\bullet$&&&&&&&&\tabularnewline
\hline
\cite{royer2017probabilistic}&&$\bullet$&&$\bullet$&&&&&&\tabularnewline
\hline
\cite{kumar2021colorization}&&$\bullet$&&&&&&&&\tabularnewline
\hline
\multicolumn{11}{|c|}{Object aware}\tabularnewline
\hline
\cite{su2020instance}&&$\bullet$&$\bullet$&&& &$\bullet$&&& \tabularnewline
\hline
\cite{pucci2021collaborative}&&$\bullet$&$\bullet$&&& &$\bullet$&&& \tabularnewline
\hline
\cite{kong2021adversarial}&&&&&&$\bullet$&&&& \tabularnewline
\hline
\multicolumn{11}{|c|}{Survey}\tabularnewline
\hline
\cite{Gu_2019_CVPR_Workshops}&&&&&$\bullet$&&&&&\tabularnewline
\hline
\cite{anwar2020image}&& &&&& &&&&{\scriptsize own Natural-Color Dataset}\tabularnewline
\hline
\end{tabular}}
\caption{Datasets used in the literature for training or testing}\label{table:datasets}
\end{table}

\section{Proposed Colorization Framework}
\label{sec:framework}

In this section, we present the framework that we will use for evaluating the influence of color spaces on image colorization results.
First, we detail the architecture, and second, the dataset used for training and testing. 

Note that the same architecture and training procedure is used in the chapter \textit{Analysis of Different Losses for Deep Learning Image Colorization} of this handbook.

\subsection{Detailed Architecture}

\begin{figure}[t]
\centering
\includegraphics[width=\linewidth]{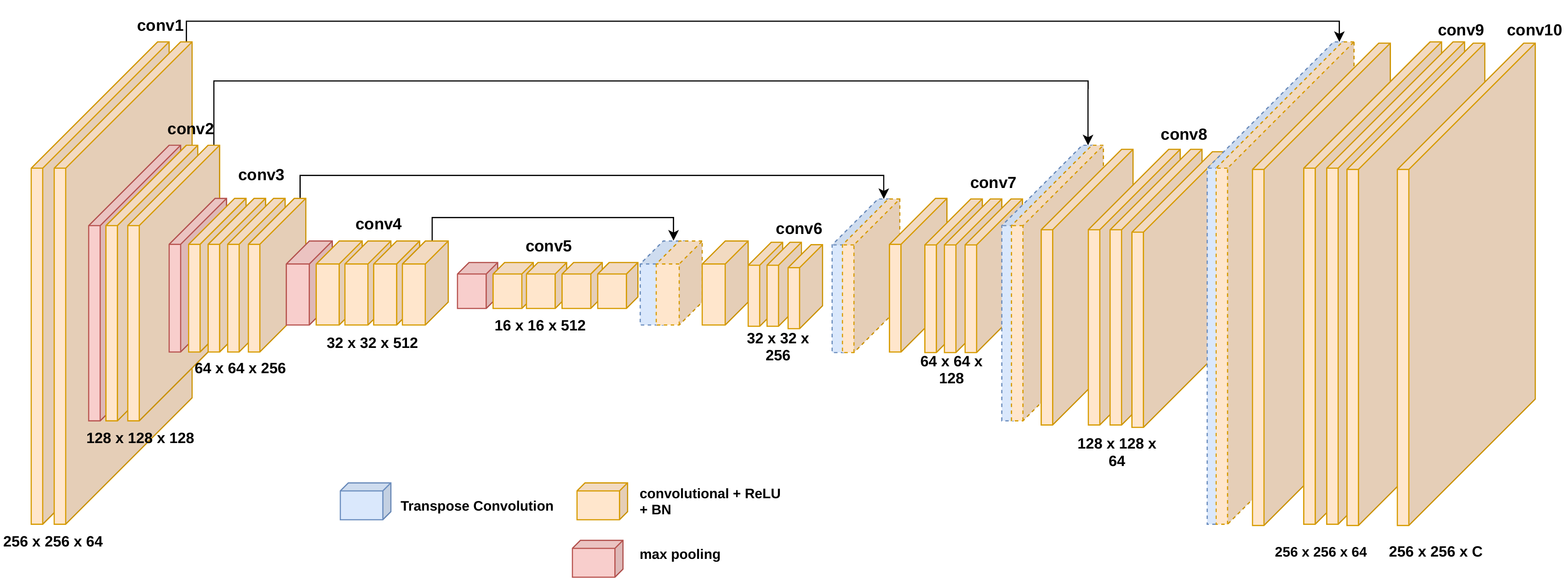}
\caption{Summary of the baseline U-Net architecture used in our experiments. It outputs a $256\times256\times C$ image, where $C$ stands for the number of channels, being equal to $2$ when estimating the missing chrominance channels and to $3$ when estimating the RGB components.}
\label{fig:unet-architecture}
\end{figure}

The architecture used in our experiments is an encoder-decoder U-Net deep network composed of five stages (see Figure~\ref{fig:unet-architecture}).
All convolutional blocks are composed of two 2D convolutional layers with $3 \times 3$ kernels, each one followed by 2D batch normalization (BN) and a ReLU activation.
For the encoder, downsampling is done with max pooling layers after each convolutional block.
After each downsampling, the number of filters is doubled in the following block.
For the decoder, upsampling is done with 2D transpose convolutions ($4 \times 4$ kernels with stride 2).
At a given stage, the corresponding encoder and decoder blocks are linked with skip connections: feature maps from the encoder are concatenated with the ones from the corresponding upsampling path and fused using $1 \times 1$ convolutions.
More details can be found in Table \ref{tab:arch-details}.
The encoder architecture is identical to the CNN part of a VGG-19 network~\cite{simonyan2017very}.
It allows us to start from pretrained weights initially used for ImageNet classification.
Moreover, the encoder architecture choice was motivated by the fact that most deep learning-based approaches use a VGG-type architecture to generate the missing chrominances.

\begin{table}[t]
\centering
\setlength{\tabcolsep}{0.4cm}
\begin{tabular}{lc}
\hline
\textbf{Layer type}           & \textbf{Output resolution} \\ \hline
Input                         & 3 x H x W                             \\
Conv1 + Max-pooling           & 64 x H/2 x W/2                        \\
Conv2 + Max-pooling           & 128 x H/4 x W/4                       \\
Conv3 + Max-pooling           & 256 x H/8 x W/8                       \\
Conv4 + Max-pooling           & 512 x H/16 x W/16                     \\
Conv5 + Conv. Transpose (I)   & 512 x H/8 x W/8                       \\
Conv6 + Conv. Transpose (II)  & 256 x H/4 x W/4                       \\
Conv7 + Conv. Transpose (III) & 128 x H/2 x W/2                       \\
Conv8 + Conv. Transpose (IV)  & 64 x H x W                            \\
Conv9                         & 64 x H x W                            \\
Conv10                        & C x H x W                             \\ \hline
\end{tabular}
\caption{Detailed architecture and output resolution for each block.}
\label{tab:arch-details}
\end{table}

The training settings are described as follows:
\begin{itemize}
\item Optimizer: Adam
\item Learning rate: 2e-5 as in ChromaGAN~\cite{vitoria2020chromagan}.
\item Batch size: 16 images (approx. 11 GB RAM usage on Nvidia Titan V).
\item All images are resized to $256 \times 256$ for training which enable using batches.
In practice, to keep the aspect ratio, the image is resized such that the smallest dimension matches $256$. If the other dimension remains larger than $256$, we then apply a random crop to obtain a square image. Note that the random crop is performed using the same seed for all trainings.
\end{itemize}

When generating images, it is crucial to remain in the range of acceptable values of color spaces.
In particular, we must ensure that the final image takes values between $0$ and $255$.
In our implementation, we use simple clipping on final RGB values.
Other strategy are sometimes considered as in \cite{Iizuka2016} where the a*b* components are globally normalized so they lie in the [0,1] range of the Sigmoid transfer function.

\subsection{Training and Testing Images}

Throughout our experiments we use the COCO dataset \cite{lin2014microsoft}, containing various natural images of different sizes.
COCO is divided into three sets that approximately contain 118k, 5k and 40k images that, respectively, correspond to the training, validation and test sets.
Note that we carefully remove all grayscale images, which represent around $3\%$ of the overall amount of each set.
Although larger datasets such as ImageNet have been regularly used in the literature, COCO offers a sufficient number and a good variety of images so we can efficiently train and compare numerous models.

\section{Learning Strategy for Different Color Spaces}
\label{sec:learning}

\begin{figure}[t!]
\centering
\includegraphics[width=0.59\linewidth]{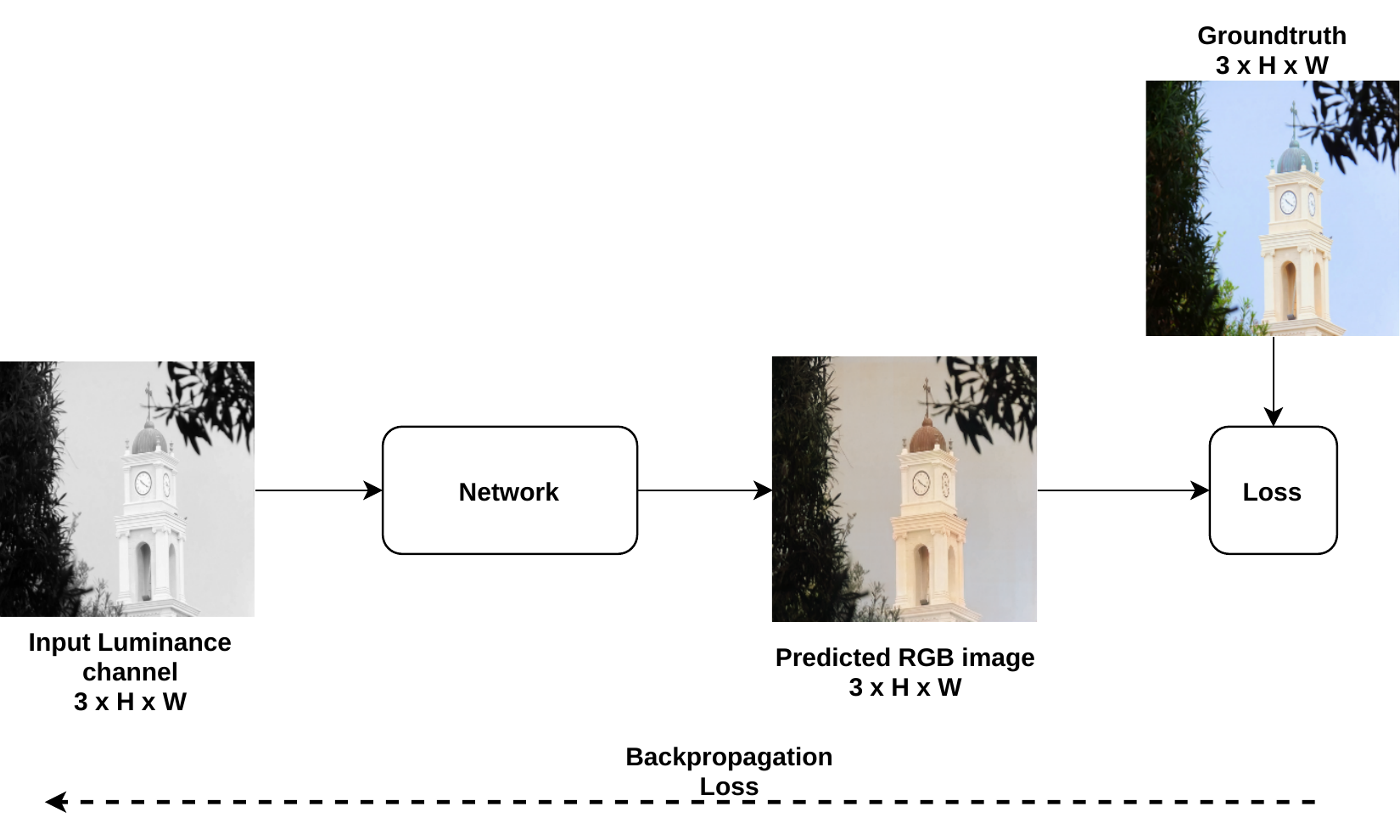}\\%
{a. Learning strategy directly predicting the RGB colors.}\\%
\includegraphics[width=0.59\linewidth]{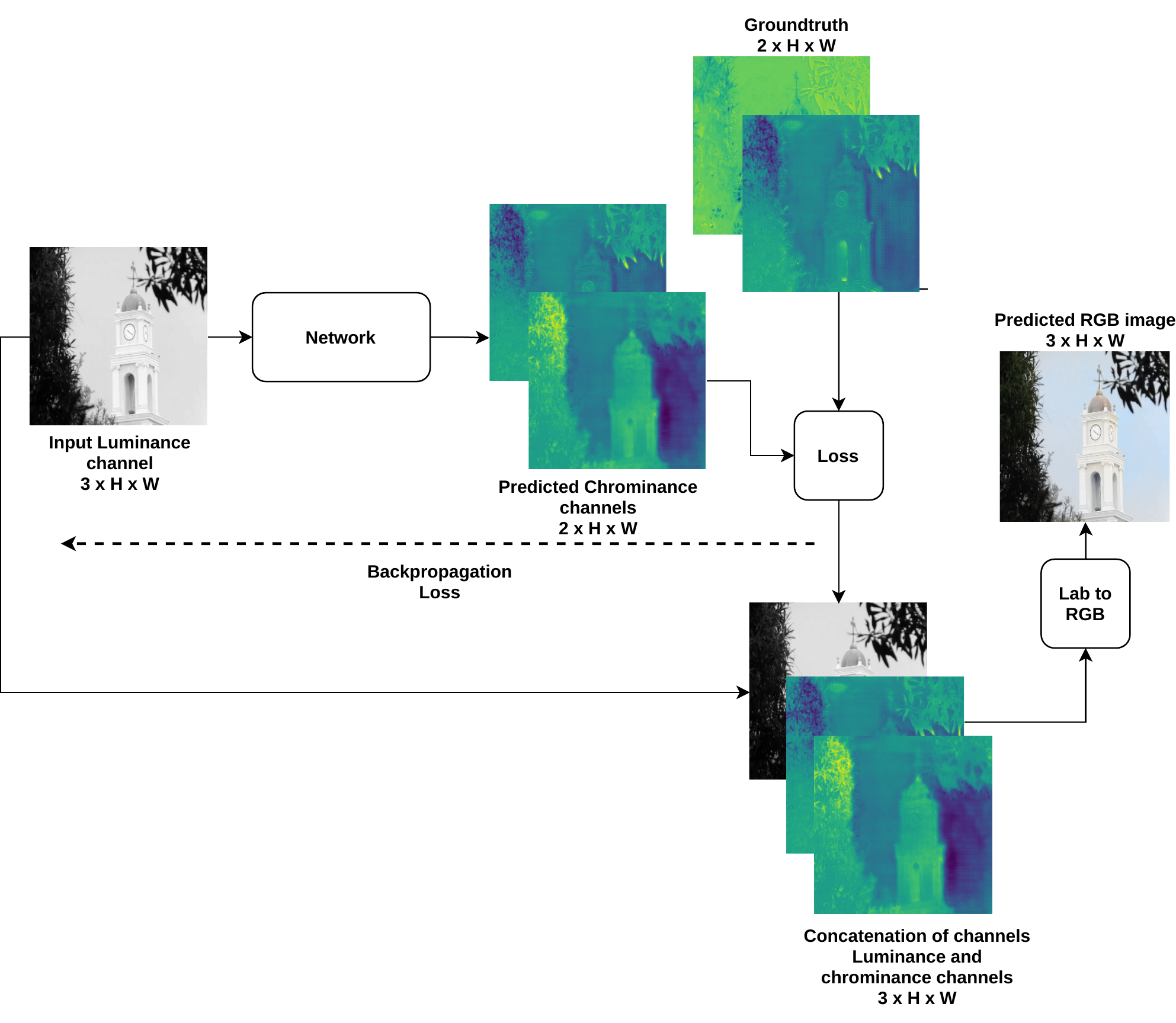}\\%
{b. Learning strategy predicting the two chrominance channels.}\\%
\includegraphics[width=0.59\linewidth]{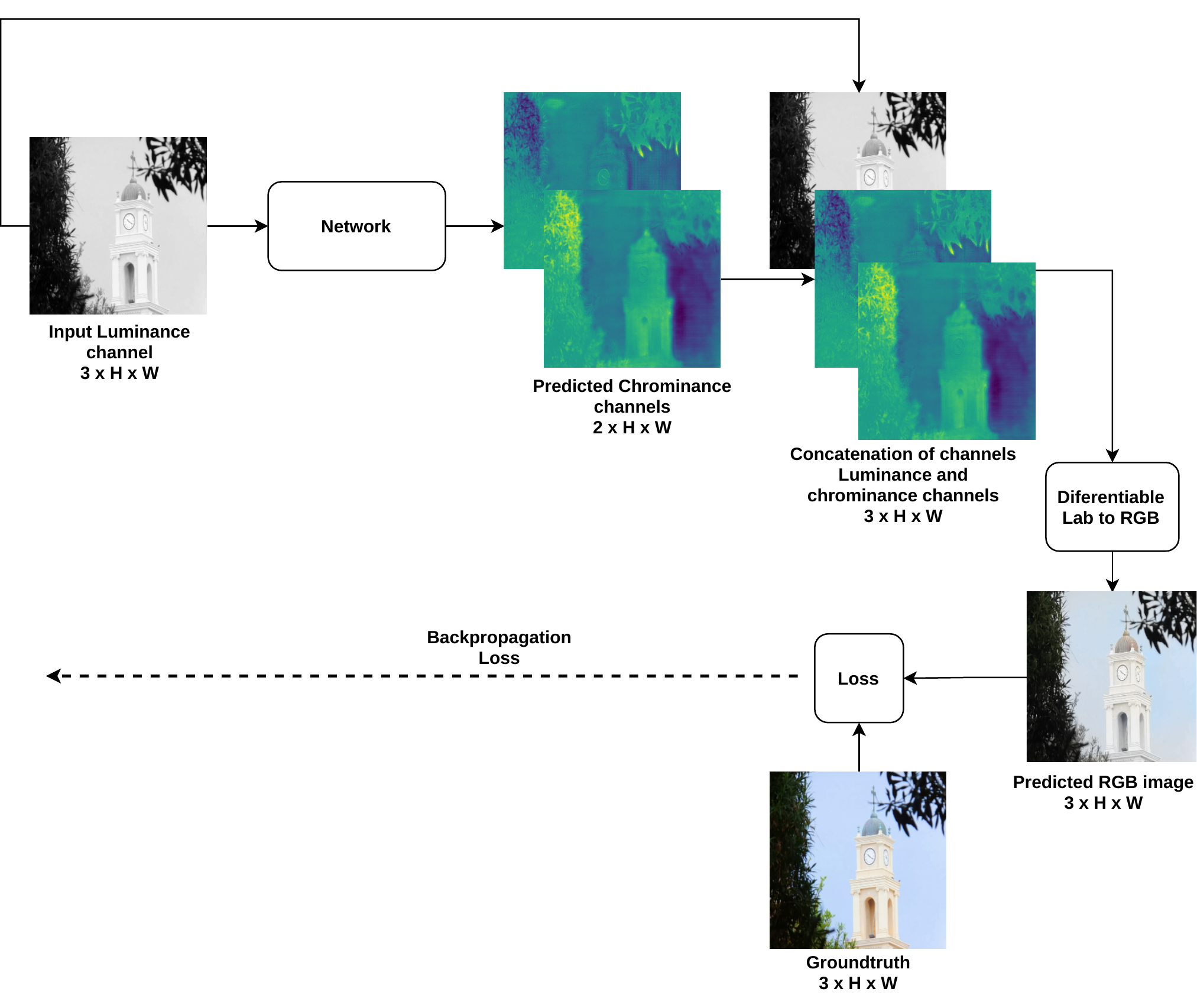}\\%
{c. Learning strategy predicting the two chrominance channels, then converting to RGB.}
\caption{Illustration of the different learning strategies for our proposed framework.}
\label{fig:strategy}
\end{figure}

The goal of the whole colorization process is to generate RGB images that look visually natural. 
When training on different color spaces, one must decide which color space is used to compute losses and when is the conversion back to RGB performed.
In this chapter, we propose to experiment with three learning strategies to compare RGB, YUV, and Lab color spaces (see Figure~\ref{fig:strategy}):
\begin{itemize}
    \item \textit{RGB}: in this case, the network takes as input a grayscale image $L$ and directly estimates a  three-channels RGB image of size $256\times 256\times 3$. The loss is done directly in the RGB color space. This strategy is illustrated in Figure~\ref{fig:strategy}a.

    \item \textit{YUV and Lab Luminance/chrominance}: in this case, the network takes as input a grayscale image considered as the luminance ($L$ for Lab, $Y$ for YUV) and outputs two chrominance channels ($a$, $b$ or $U$, $V$). The loss compares the output with the corresponding chrominance channels of the ground truth image converted to the luminance/chrominance space. After concatenating the initial luminance channel to the inferred chrominances, the image is converted back to RGB for visualization purposes. This strategy is illustrated in Figure~\ref{fig:strategy}b.

    \item \textit{LabRGB}: as in the previous case, the network takes as input the luminance and estimates the corresponding two chrominance channels. After concatenating with the corresponding luminance channel, they are converted to the RGB colorspace and the loss is computed directly there. Notice, that in this last case, as the loss is computed on RGB colorspace, the conversion must be done in a way that is differentiable to be able to compute the gradient and allow the back-propagation step. We perform the color conversion using the color module in the \texttt{Kornia} library. Kornia~\cite{riba2020kornia} is a differentiable library that consists of a set of routines and differentiable modules to solve generic computer vision problems.  It allows classical computer vision tasks to be integrated into deep learning models. Computing the loss on RGB images instead of chrominance ones enables to ensure images are similar to ground truth after the clipping operation needed to fit into the RGB cube. This strategy is illustrated in Figure \ref{fig:strategy}c.
\end{itemize}

\textit{Remark:} During training, all images are resized to 256$\times$ 256. One advantage of using luminance/chrominance spaces is that only chrominance channels are resized. It is therefore possible to keep the original content of the luminance channels without manipulating it with the resizing steps.

\section{Analysis of the Influence of Color Spaces }
\label{sec:colorinfluence}

This section presents quantitative and qualitative results obtained with the three strategies discussed above.
For this analysis, we have considered, as loss function, the L2 loss and the VGG-based LPIPS which was introduced in~\cite{ding2021comparison} as a generalization of the feature loss~\cite{johnson2016perceptual}.
These loss functions are defined hereafter.

\textbf{MSE or squared L2 loss.}
The L2 loss, between two functions $u$ and $v$ defined on $\Omega$ and with values in $\mathbb{R}^C$, $C\in\mathbb{N}$, is defined as the squared L2 loss of their difference.
That is,
\begin{equation}
\mathrm{MSE}(u,v)=\|u-v\|_{L^2(\Omega;\mathbb{R}^C)}^{2}=\int_{\Omega}\|u(x)-v(x)\|_2^2 dx,
\end{equation}
where $\|\cdot\|_2$ denotes the Euclidean norm in $\mathbb{R}^C$. In the discrete setting, it is equal to the sum of the square differences between the image values, that is,
\begin{equation}
\mathrm{MSE}(u,v)=\sum_{i=1}^M\sum_{j=1}^{N}\sum_{k=1}^C (u_{i,j,k}-v_{i,j,k})^2.
\end{equation}

\noindent
\textbf{Feature Loss.} The feature reconstruction loss \cite{gatys2015neural,johnson2016perceptual} is a perceptual loss that encourages images to have similar feature representations as the ones computed by a pretrained network, denoted here by $\Phi$.
Let $\Phi_l({u})$ be the activation of the $l$-th layer of the network $\Phi$ when processing the image ${u}$; if $l$ is a convolutional layer, then $\Phi_l({u})$ will be a feature map of size $C_l \times W_l \times H_l$.
The \emph{feature reconstruction} loss is the normalized squared Euclidean distance between feature representations, that is,
\begin{equation}
\mathcal{L}_{\text{feat}}^l(u,v) = \frac{1}{C_l W_l H_l}\left\lVert \Phi_{l}(u) - \Phi_{l}(v) \right\rVert_2^2 .
\label{eq:featureReconstruction}
\end{equation}
It penalizes the output reconstructed image when it deviates in feature content from the target.

\noindent
\textbf{LPIPS.} LPIPS~\cite{Zhang_2018_CVPR} computes a weighted L2 distance between deep features of a pair of images $u$ and $v$:
\begin{equation}\label{eq:LPIPSeq}
\rm{LPIPS}(u,v) = \sum_l\frac{1}{H_lW_l}\sum_{i=1}^{H_l}\sum_{j=1}^{W_l}\|\omega_l\odot (\Phi_l(u)_{i,j} - \Phi_l(v)_{i,j} )\|_2^2 ,
\end{equation}
where $H_l$ (resp. $W_l$) is the height (resp. the width) of feature map $\Phi_l$ at layer $l$ and $\omega_l$ are weights for each features. Note that features are unit-normalized in the channel dimension. We will denote VGG-based LPIPS when feature maps $\Phi_l$ are taken from a VGG network.

Note that to compute the VGG-based LPIPS loss, the output colorization always has to be converted to RGB, even for YUV and Lab color spaces (as in Figure~\ref{fig:strategy}(c)), because this loss is computed with a pre-trained VGG expecting RGB images as input.
Since VGG-based LPIPS is computed on RGB images, the two strategies \textit{Lab} and \textit{LabRGB} are the same.
For more details on the various losses usually used in colorization, we refer the reader to the chapter \textit{Analysis of Different Losses for Deep Learning Image Colorization}.
Our experiments have shown that same conclusions can be drawn with other losses.\\ For testing, we apply the network to images at their original resolution, while training is done on batches of square $256\times 256$ images.

\subsection{Quantitative Evaluation}

There is no standard protocol for quantitative evaluation of automatic colorization methods.
We refer the reader to the chapter \textit{Analysis of Different Losses for Deep Learning Image Colorization} for a detailed survey of quantitative evaluation methods used in image colorization literature, and analysis of correlation between losses versus type of evaluation metrics used.
We choose to rely on the more generally used and more recent ones: L1 (MAE), L2 (MSE), PSNR, SSIM~\cite{wang2004image}, LPIPS~\cite{Zhang_2018_CVPR} and FID (Fréchet Inception Distance)~\cite{dowson1982frechet}, which are defined hereafter.

\textbf{MAE or L1 loss with $l^1$-coupling.}
The Mean Absolute Error is defined as the L1 loss with $l^1$-coupling, that is,
\begin{equation}\label{eq:mae}
\mathrm{MAE}(u,v)=\int_{\Omega}\|u(x)-v(x)\|_{l^1} dx=\int_{\Omega}\sum_{k=1}^C |u_k(x)-v_k(x)| dx.
\end{equation}
In the discrete setting, it coincides with the sum of the absolute differences $|u_{i,j,k}-v_{i,j,k}|$.
Some authors use a $l^2$-coupled version of it, 
\begin{equation}
\mathrm{MAE}^c(u,v)=\sum_{i=1}^M\sum_{j=1}^{N} \sqrt{\sum_{k=1}^C (u_{i,j,k}-v_{i,j,k})^2}.
\end{equation}
Both MAE and $\mathrm{MAE}^c$ losses are robust to outliers.

\textbf{PSNR.} The PSNR measures the ratio between the maximum value of a color target image $u:\Omega\rightarrow \mathbb{R}^C$ and the mean square error (MSE) between $u$ and a colorized image $v:\Omega\rightarrow \mathbb{R}^C$ with $\Omega\in\mathbb{Z}^2$ a discrete grid of size $M\times N$. That is,
\begin{align}
\rm{PSNR}(u,v) = 20&\log_{10}(\max{u})   \nonumber \\
&- 10\log_{10} \left(\frac{1}{CMN}\sum_{k=1}^C\sum_{i=1}^M\sum_{j=1}^N (u(i,j,k)-v(i,j,k))^2\right),
\end{align}
where $C=3$ when working in the RGB color space and $C=2$ in any luminance-chrominance color space as YUV, Lab and YCbCr.
The PSNR score is considered as a reconstruction measure tending to favor methods that will output results as close as possible to the ground truth image in terms of the MSE.

\textbf{SSIM.} SSIM intends to measure the perceived change in structural information between two images.
It combines three measures to compare images color~($l$), contrast~($c$) and structure~($s$):
\begin{equation}
\rm{SSIM}(u,v) = l(u,v)c(u,v)s(u,v)\\
= \frac{\left(2\mu_u\mu_v\right)+c_1}{\mu_u^2+\mu_v^2+c_1}\frac{\left(2\sigma_u\sigma_v+c_2\right)}{\sigma_u^2+\sigma_v^2+c_2}\frac{\left(\sigma_{uv}+c_3\right)}{\sigma_u\sigma_v+c_3},
\end{equation}
where $\mu_u$ (resp. $\sigma_u$) is the mean value (resp. the variance) of image $u$ values and $\sigma_{uv}$ the covariance of $u$ and $v$. $c_1$, $c_2$, $c_3$ are regularization constants that are used to stabilize the division for images with mean or standard deviation close to zero.

\textbf{FID.}
FID~\cite{heusel2017gans} is a quantitative measure used to evaluate the quality of the outputs' generative model and which aims at approximating human perceptual evaluation.
It is based on the Fréchet distance~\cite{dowson1982frechet} which measures the distance between two multivariate Gaussian distributions.
FID is computed between the feature-wise mean and covariance matrices of the features extracted from an Inception v3 neural network applied to the input images $(\mu_r,\Sigma_r)$ and those of the generated images $(\mu_g,\Sigma_g)$:
\begin{equation}
\rm{FID}\left((\mu_r,\Sigma_r),(\mu_g,\Sigma_g)\right) = \Vert \mu_r - \mu_g \Vert^2_2 + Tr(\Sigma_r + \Sigma_g - 2\Sigma_r \Sigma_g)^{1/2}.
\end{equation}

The results are presented in Table~\ref{table:Quantitativeresults}.
{In terms of these metrics, the best results are obtained with YUV color space except for L1 and Fréchet Inception Distance, even if not by much.}
{The results in Table~\ref{table:Quantitativeresults} also indicate that Lab does not outperform other color spaces when using a classic reconstruction loss (L2), while better results are obtained when using the VGG-based LPIPS.} {Thus, using a feature based reconstruction loss is better suited as was already the case in exemplar-based image colorization methods where different features for patch-based metrics were proposed for matching pixels.}
{LabRGB strategy gets the worst quantitative results based on Table~\ref{table:Quantitativeresults}. One would expect to get the "best of both" color spaces while recovering from the loss of information in the conversion process. However, this is not reflected with these particular evaluation metrics.}
The LabRGB line for VGG-based LPIPS is not included, as it would be identical to the Lab one. Also, note that the quantitative evaluation is performed on RGB images as opposed to training which is done for specific color spaces (RGB, YUV, Lab and LabRGB).

\begin{table}[t]
\centering
\setlength{\tabcolsep}{0.15cm}
\renewcommand{\arraystretch}{1.2}
\begin{tabular}{c|c|cccccc}
\hline
\textbf{Color space} & \textbf{Loss function} &  \textbf{L1} $\downarrow$  & \textbf{L2} $\downarrow$ &  \textbf{PSNR}  $\uparrow$  & \textbf{SSIM}  $\uparrow$ & \textbf{LPIPS} $\downarrow$   & \textbf{FID} $\downarrow$    \\ \hline
RGB & L2    & \textbf{0.04458} & 0.00587 & 22.3136 & \underline{0.9255} & \underline{0.1606} & \textbf{7.4223} \\
YUV & L2    & \underline{0.04469} & \textbf{0.00562} & \textbf{22.5052} &	\textbf{0.9278} & \textbf{0.1593} & \underline{7.6642} \\
Lab & L2    & 0.04488 & \underline{0.00585} & \underline{22.3283}	& 0.9250 & 0.1613 & 8.1517 \\
LabRGB & L2 & 0.04608 & 0.00589 & 22.2989 & 0.9209 & 0.1698 & 8.3413 \\
\hline
\hline
RGB & LPIPS & 0.04573 &	0.00577 & 22.3892 & \underline{0.9197} & 0.1429 & \textbf{3.0576} \\
YUV & LPIPS & \underline{0.04460} & \textbf{0.00557} & \textbf{22.5438} & 0.9097 & \textbf{0.1400} & 3.3260 \\
Lab & LPIPS & \textbf{0.04374} & \underline{0.00566} & \underline{22.4699} & \textbf{0.9228} & \underline{0.1403} & \underline{3.2221} \\
\hline
\end{tabular}
\caption{Quantitative evaluation of colorization results for different color spaces. Metrics are used to compare ground-truth to every images in the 40k test set. Best and second best results by column are in bold and underlined respectively.}
\label{table:Quantitativeresults}
\end{table}

\subsection{Qualitative Evaluation}

In this section, we qualitatively analyze the results obtained by training the network with different color spaces as explained in Section~\ref{sec:learning}.

\begin{figure}[t] \begin{center}
\begin{tabular}{ccccc}
    \includegraphics[width=0.2\textwidth]{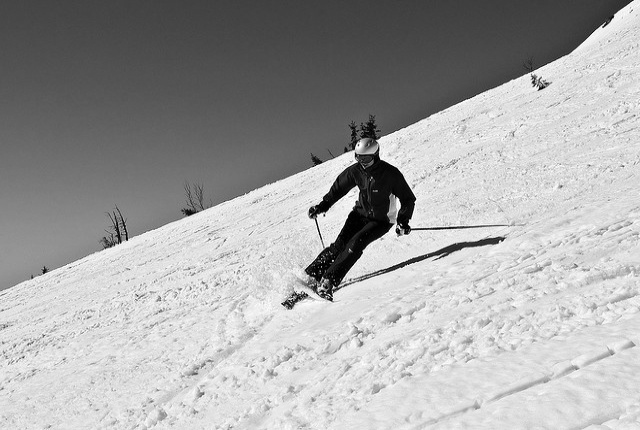}
& 
    \includegraphics[width=0.2\textwidth]{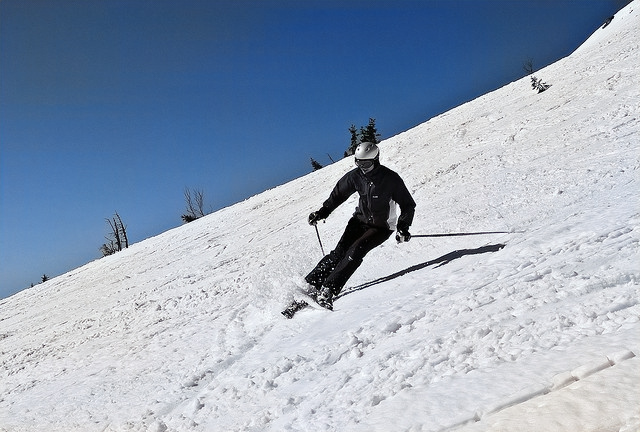}
& 
    \includegraphics[width=0.2\textwidth]{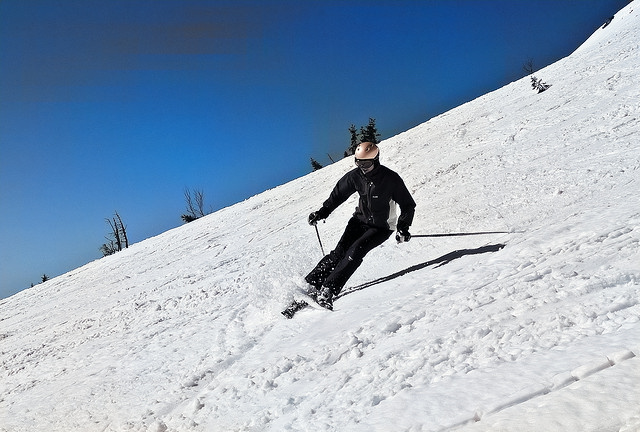}
& 
    \includegraphics[width=0.2\textwidth]{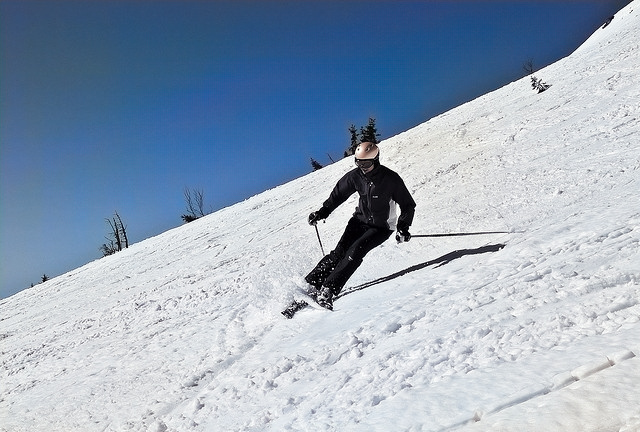}
& 
    \includegraphics[width=0.2\textwidth]{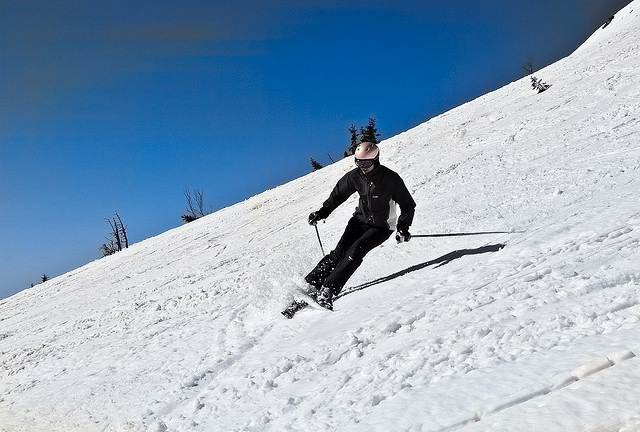}
\\
    \includegraphics[width=0.2\textwidth]{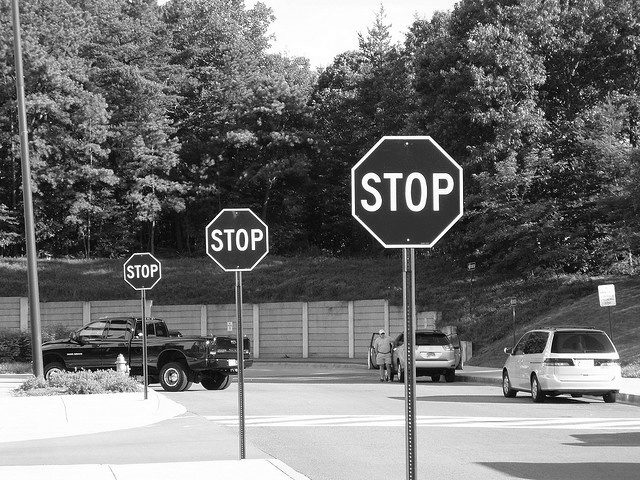}
& 
    \includegraphics[width=0.2\textwidth]{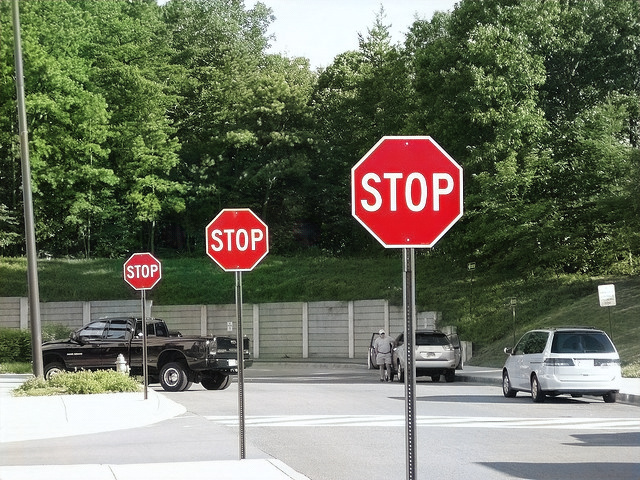}
& 
    \includegraphics[width=0.2\textwidth]{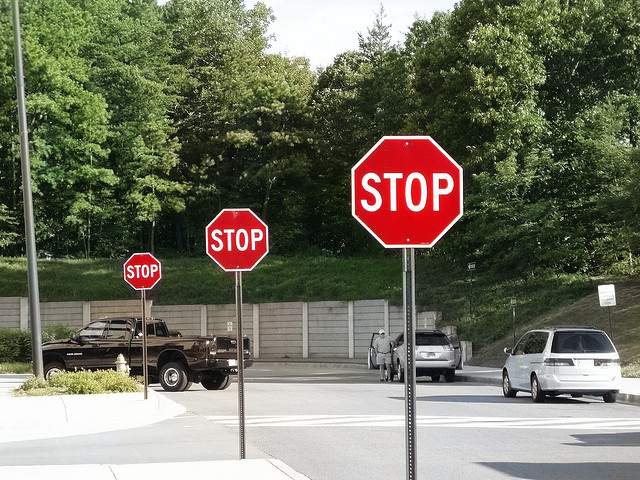}
& 
    \includegraphics[width=0.2\textwidth]{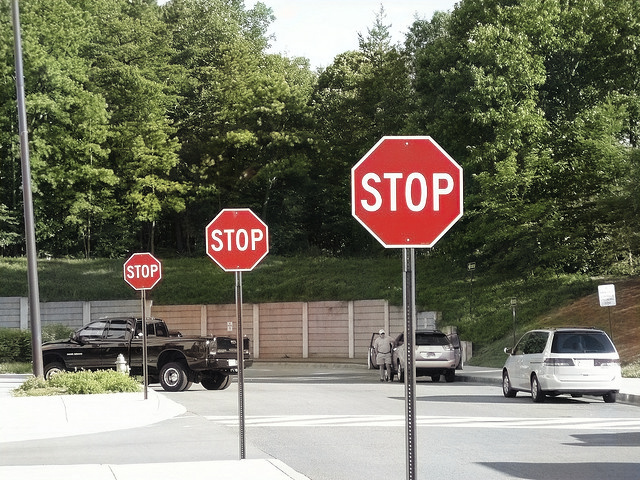}
& 
    \includegraphics[width=0.2\textwidth]{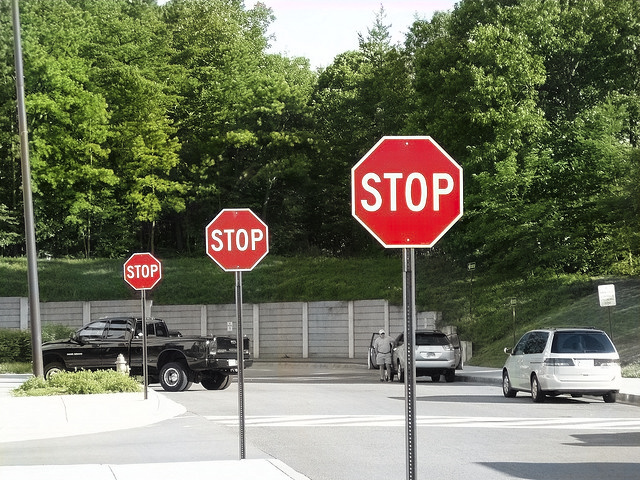}
\\
    \includegraphics[width=0.2\textwidth]{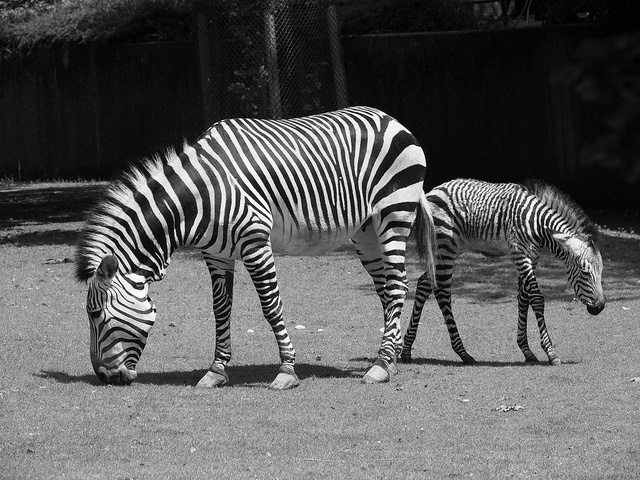}
& 
    \includegraphics[width=0.2\textwidth]{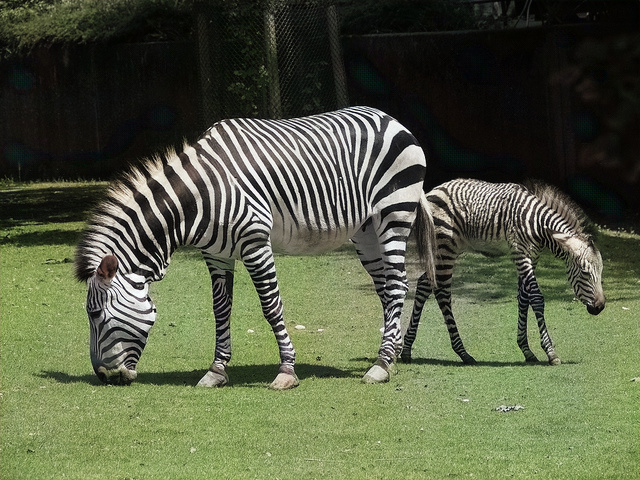}
& 
    \includegraphics[width=0.2\textwidth]{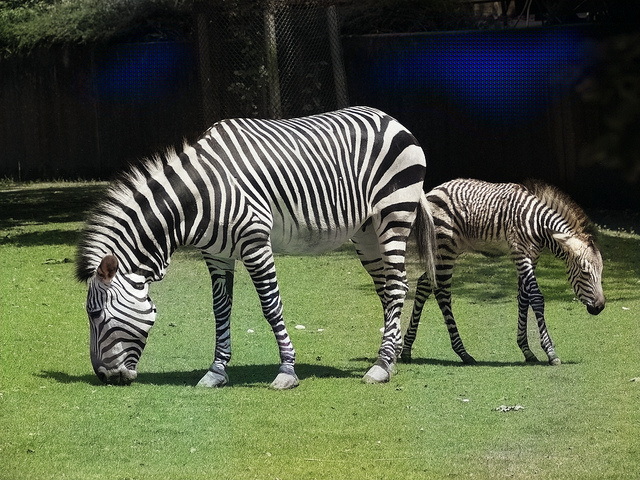}
& 
    \includegraphics[width=0.2\textwidth]{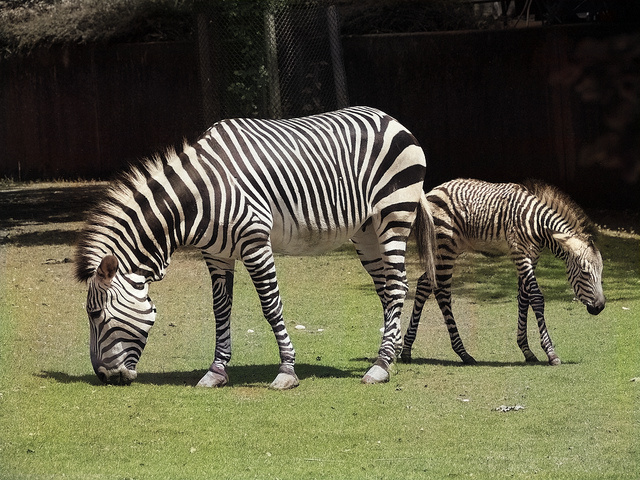}
& 
    \includegraphics[width=0.2\textwidth]{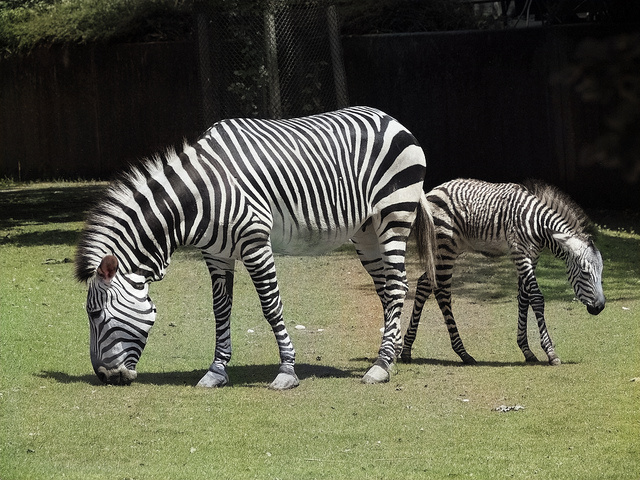}
\\
 Input   & 
 RGB-L2  & 
 YUV-L2  & 
 Lab-L2  & 
LabRGB-L2
\\
 \end{tabular}
 \begin{tabular}{ccc}
    \includegraphics[width=0.2\textwidth]{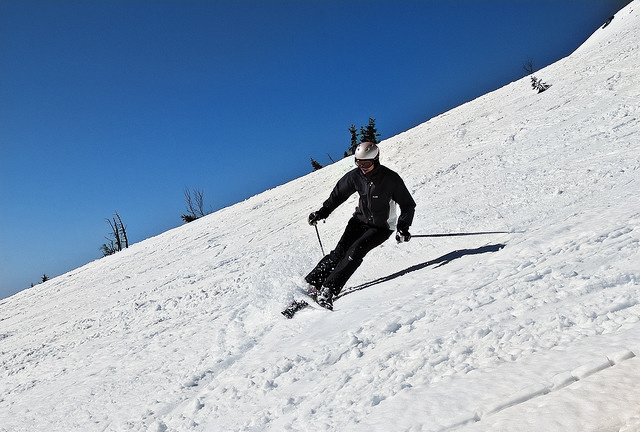}
& 
    \includegraphics[width=0.2\textwidth]{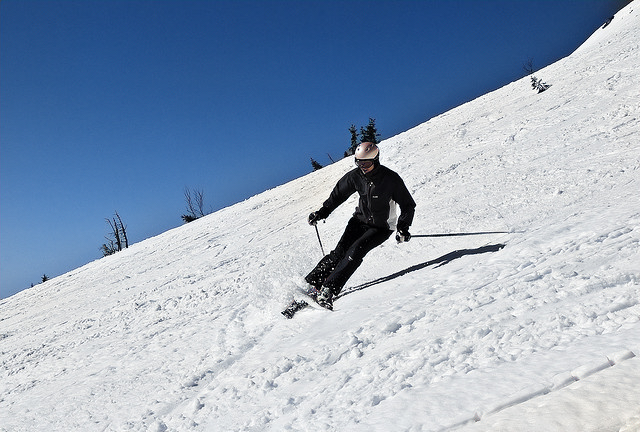}
& 
    \includegraphics[width=0.2\textwidth]{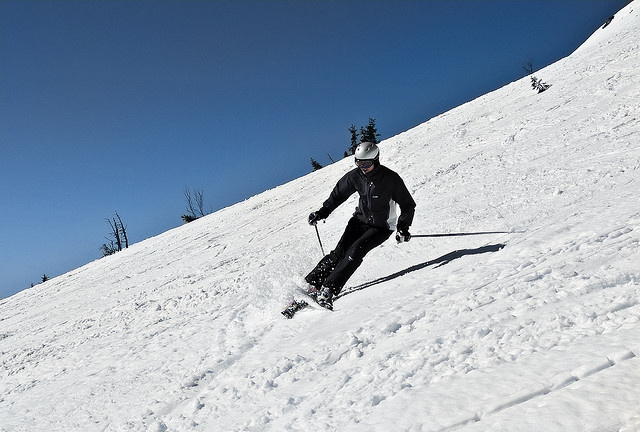}
\\
    \includegraphics[width=0.2\textwidth]{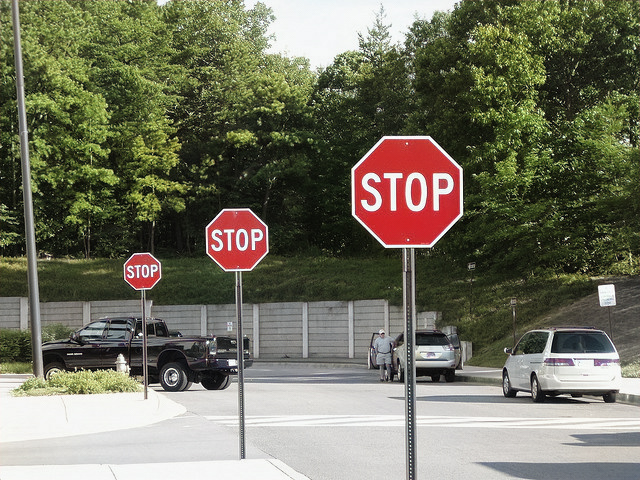}
& 
    \includegraphics[width=0.2\textwidth]{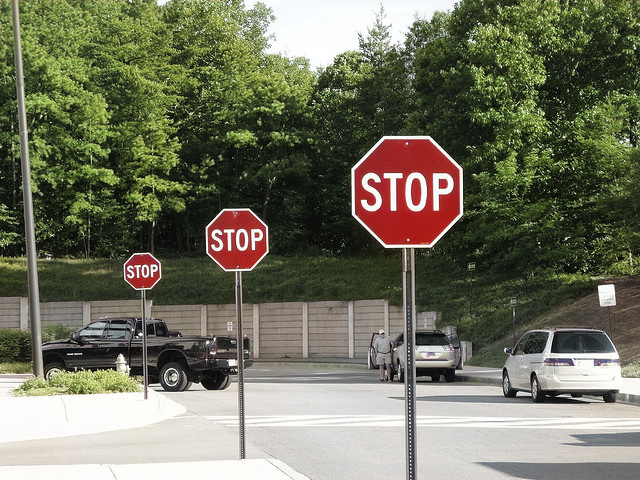}
& 
    \includegraphics[width=0.2\textwidth]{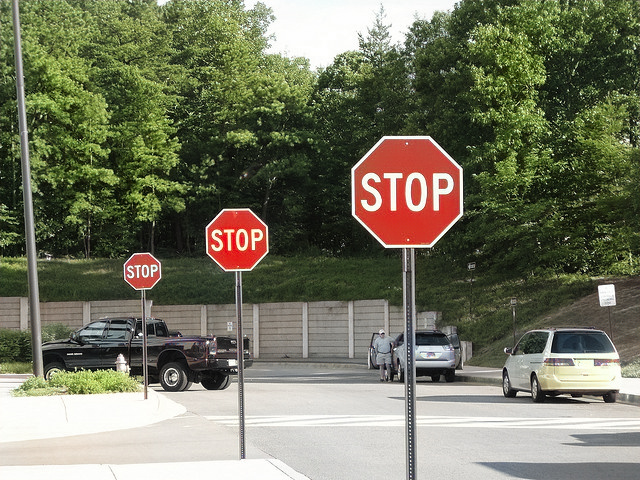}
\\
    \includegraphics[width=0.2\textwidth]{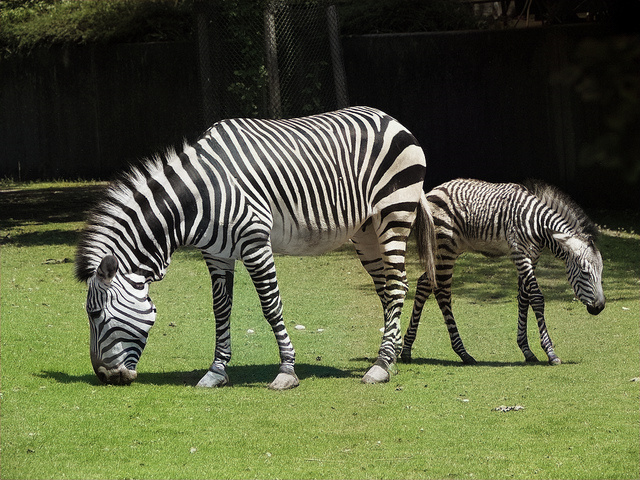}
& 
    \includegraphics[width=0.2\textwidth]{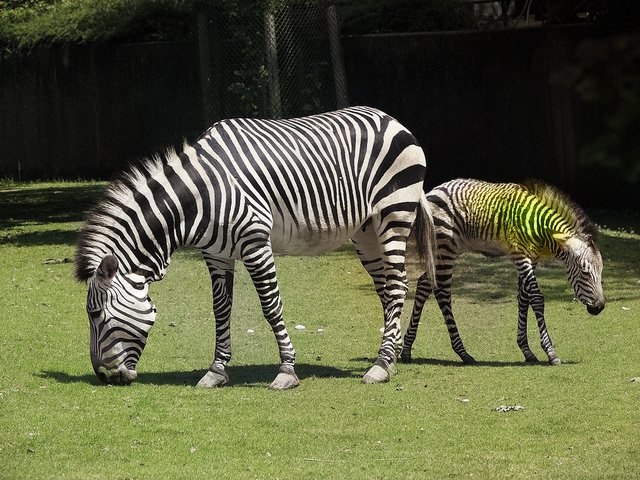}
& 
    \includegraphics[width=0.2\textwidth]{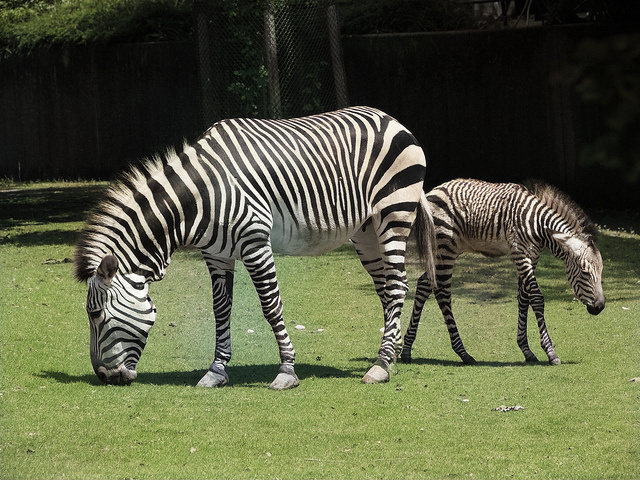}
\\
 RGB-LPIPS & 
 YUV-LPIPS  & 
Lab-LPIPS
\\
 \end{tabular}
    \caption{Colorization results with different color spaces on images that contain objects, have strong structures and that have been seen many times in the training set. The three first rows are with L2 loss and the three last ones with VGG-based LPIPS. }\label{strongstucturesL2LPIPS}
 \end{center} \end{figure}

Figure~\ref{strongstucturesL2LPIPS} shows results on images and objects (here person skiing, stop sign and zebra) with strong contours that were highly present in the training set.
The colorization of these images is really impressive for any color space.
Nevertheless, YUV has the tendency to sometimes create artifacts that are not predictable.
This is visible with the blue stain in the YUV-L2 zebra and the yellow spot in the YUV-LPIPS zebra.
One can also notice that the overall colorization tends to be more homogeneous with LabRGB-L2 than with Lab-L2 as can be seen for instance on the wall behind the stop signs, the grass and tree leafs in the zebra image which suggests that it might be better to compute losses over RGB images.
A similar remark is valid for the VGG-based LPIPS results as can be seen for instance in the homogeneous colorization of the sky in the person skiing image where the loss is again computed over the RGB image.
This indicates that there could be an additional influence on the results when using VGG-based LPIPS given that the predicted color image is converted back to RGB before backpropagation.

\begin{figure}[t]
\begin{center}
\begin{tabular}{ccccc}
    \includegraphics[width=0.2\textwidth]{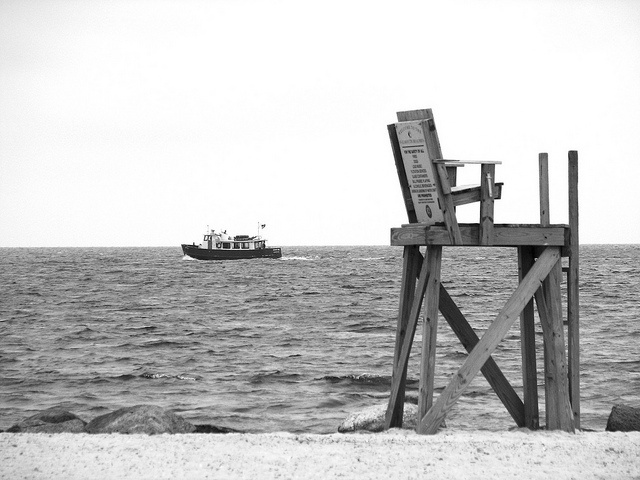}
& 
    \includegraphics[width=0.2\textwidth]{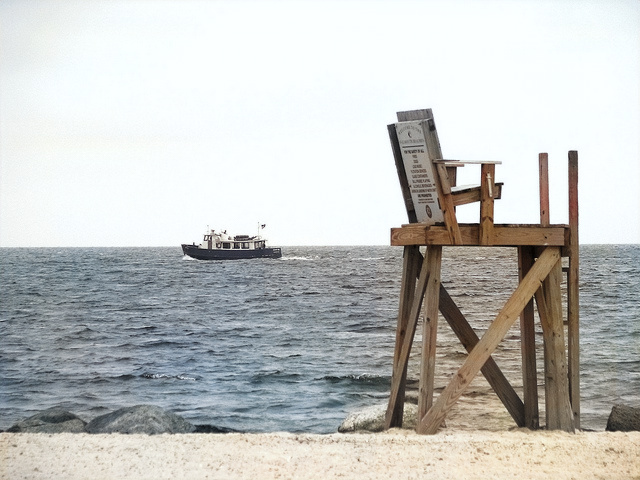}
& 
    \includegraphics[width=0.2\textwidth]{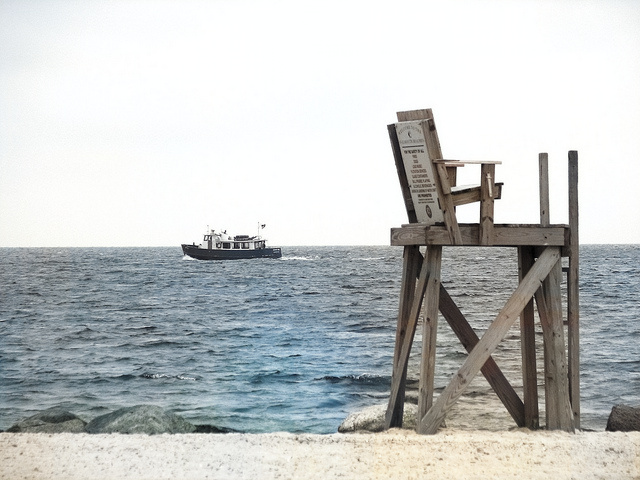}
& 
    \includegraphics[width=0.2\textwidth]{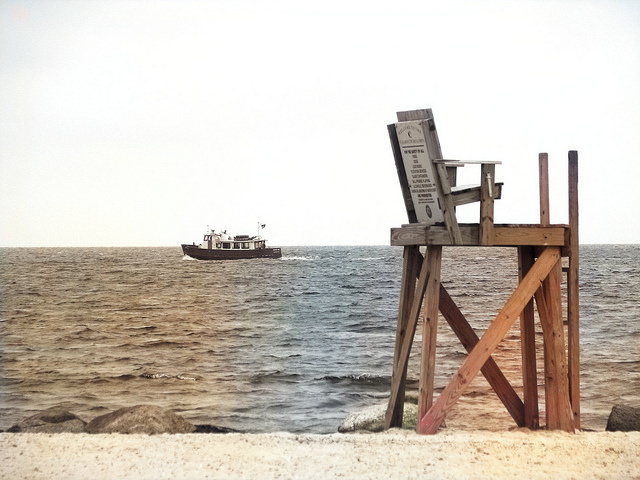}
& 
    \includegraphics[width=0.2\textwidth]{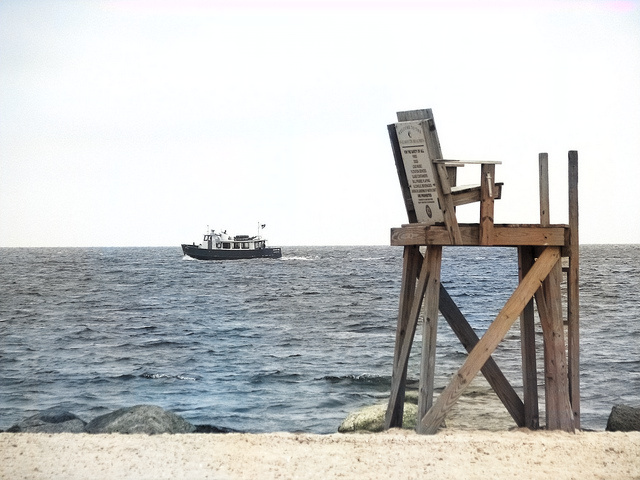}
\\
    \includegraphics[width=0.2\textwidth]{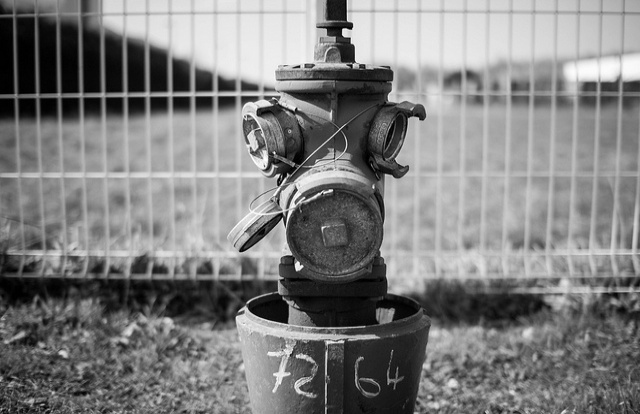}
& 
    \includegraphics[width=0.2\textwidth]{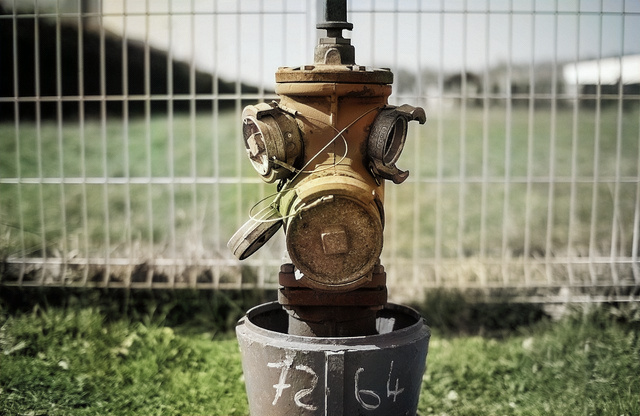}
& 
    \includegraphics[width=0.2\textwidth]{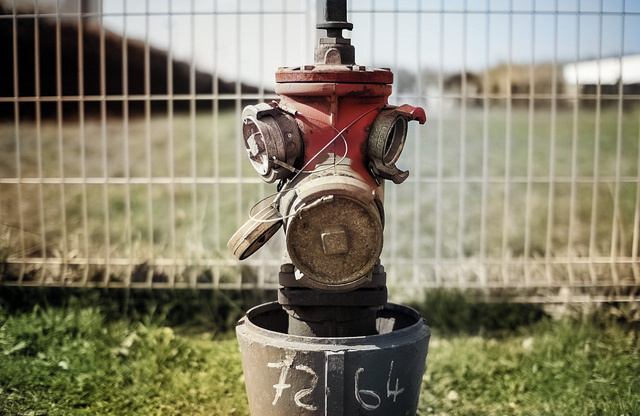}
& 
    \includegraphics[width=0.2\textwidth]{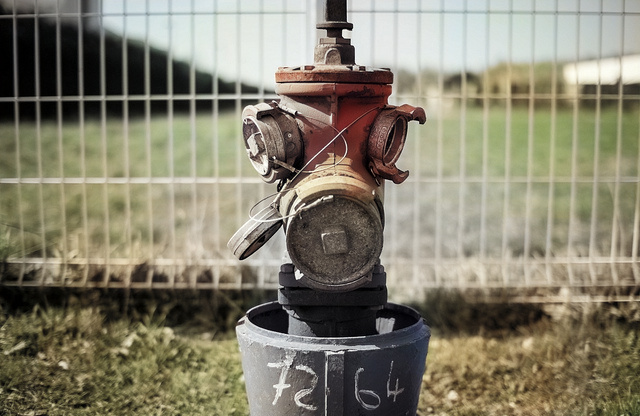}
& 
    \includegraphics[width=0.2\textwidth]{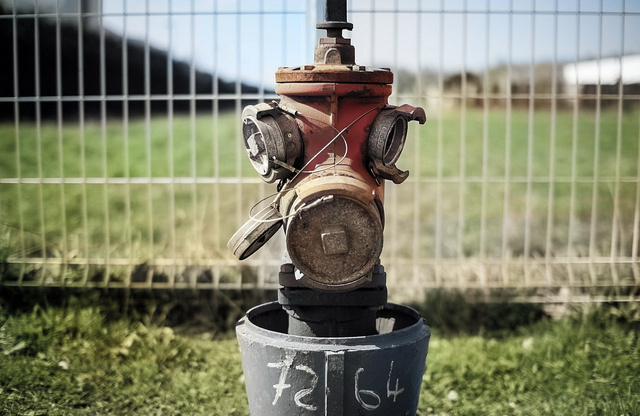}
\\
Input   &  
RGB-L2  & 
YUV-L2  & 
Lab-L2  & 
LabRGB-L2
\\
\end{tabular}
\begin{tabular}{ccc}
    \includegraphics[width=0.2\textwidth]{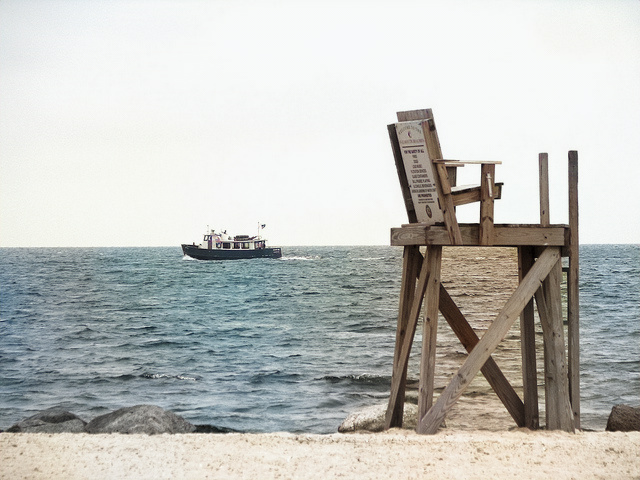}
& 
    \includegraphics[width=0.2\textwidth]{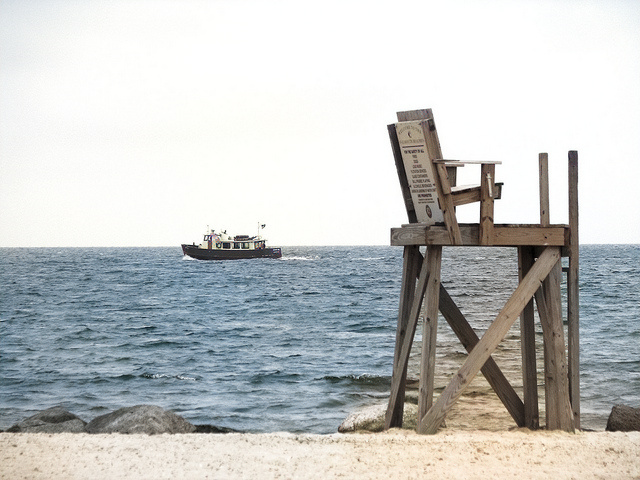}
& 
    \includegraphics[width=0.2\textwidth]{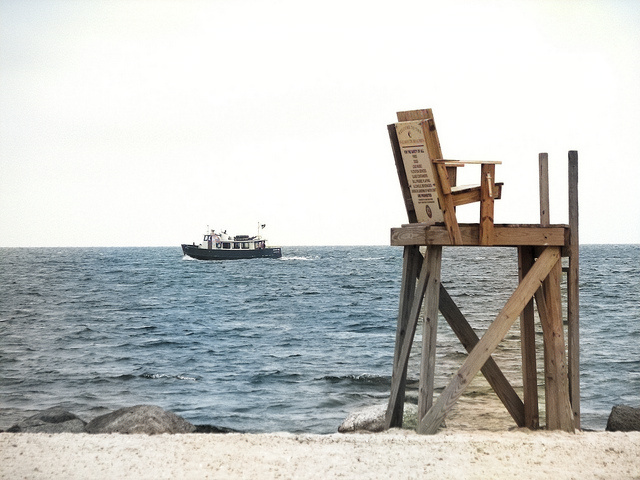}
\\
    \includegraphics[width=0.2\textwidth]{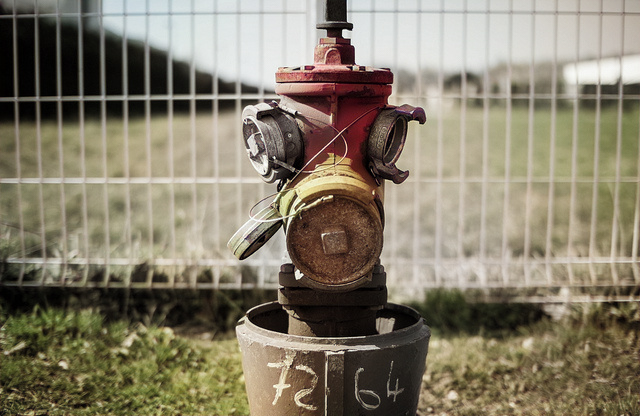}
& 
    \includegraphics[width=0.2\textwidth]{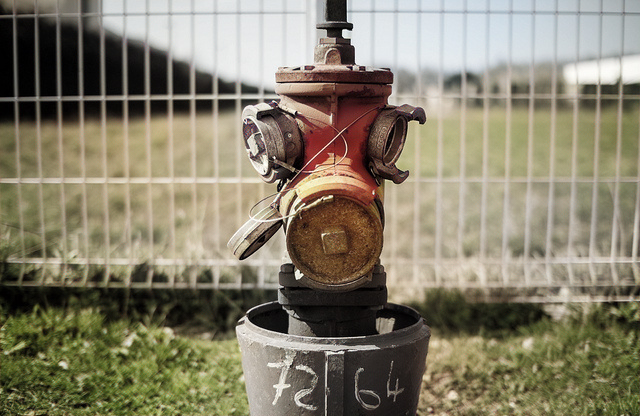}
& 
    \includegraphics[width=0.2\textwidth]{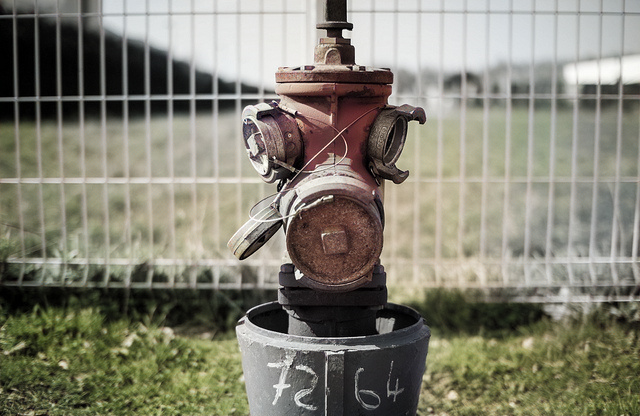}
\\
RGB-LPIPS & 
YUV-LPIPS & 
Lab-LPIPS
\\
\end{tabular}
\caption{Colorization results with different color spaces on images that exhibit strong structures that may lead to inconsistent spatial colors. The two first rows are with L2 loss and the two last one with VGG-based LPIPS.}
\label{globalconsistencyL2}
\end{center}
\end{figure}

Figure~\ref{globalconsistencyL2} presents results on images where the final colorization is not consistent over the whole image.
On the first row, the color of the water is stopped by the chair legs.
On the second row the color of the grass and the sky are not always similar on both side of the hydrant.
LabRGB seems to reduce this effect.
This happens when strong contours seem to stop the colorization and is independent on the color space.
Global coherency can only be obtained if the receptive field is large enough and that self-similarities present in natural images is preserved.
These results highlight that efforts must be put on the design of architectures that would impose these constraints.

\begin{figure}[t] \begin{center}
\begin{tabular}{ccccc}
    \includegraphics[width=0.2\textwidth]{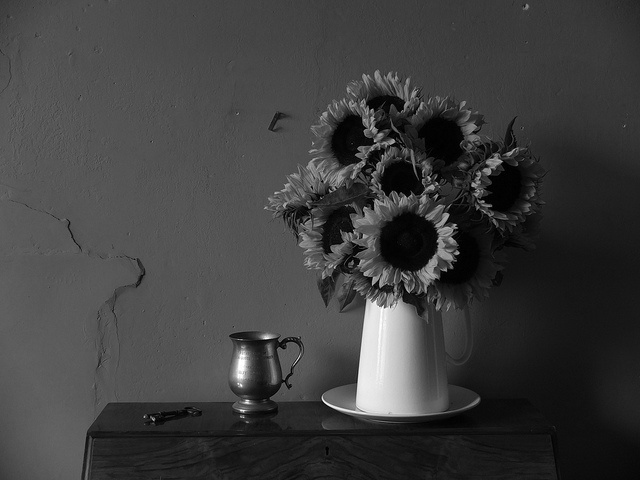}
&
    \includegraphics[width=0.2\textwidth]{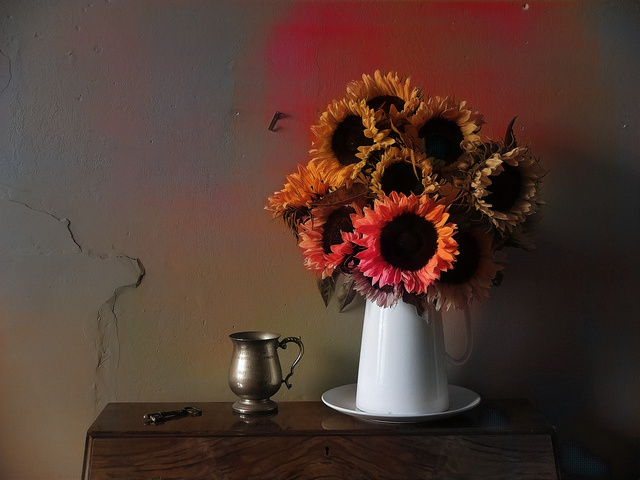}
&
    \includegraphics[width=0.2\textwidth]{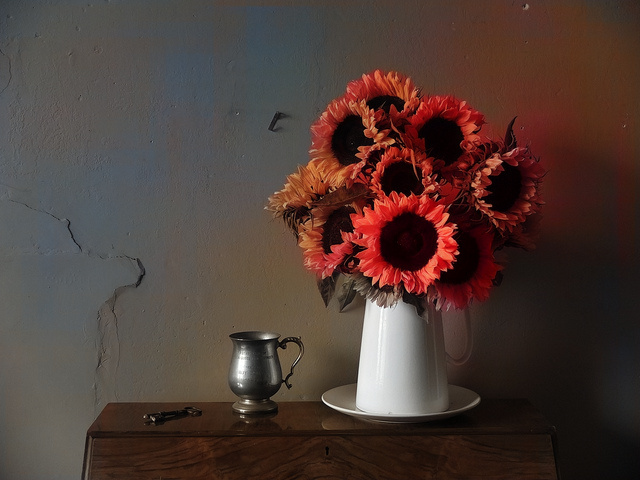}
&
    \includegraphics[width=0.2\textwidth]{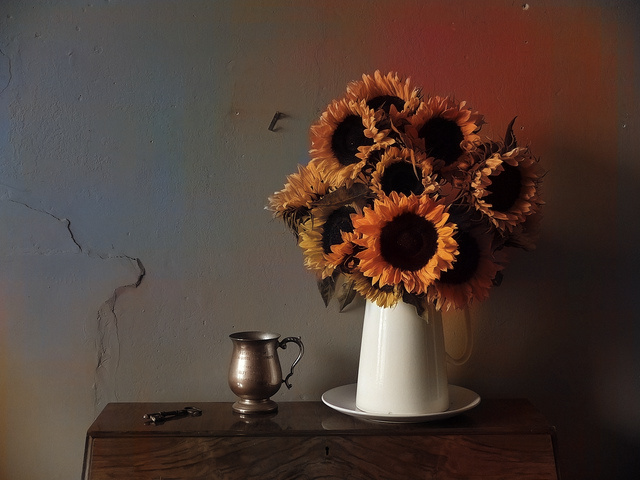}
&
    \includegraphics[width=0.2\textwidth]{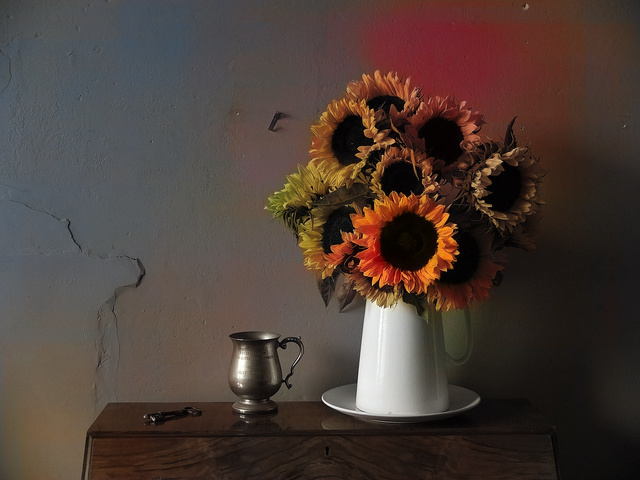}
\\
    \includegraphics[width=0.2\textwidth]{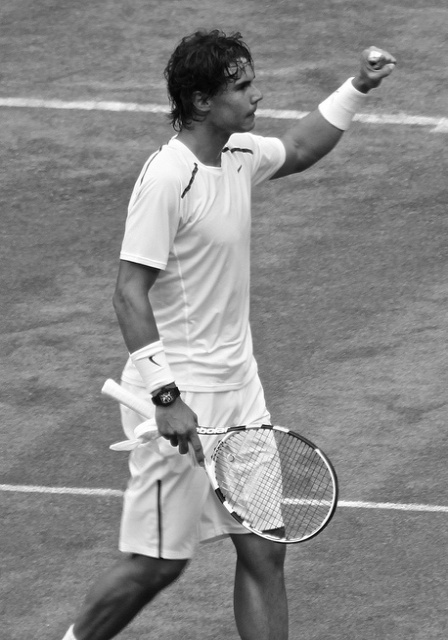}
&
    \includegraphics[width=0.2\textwidth]{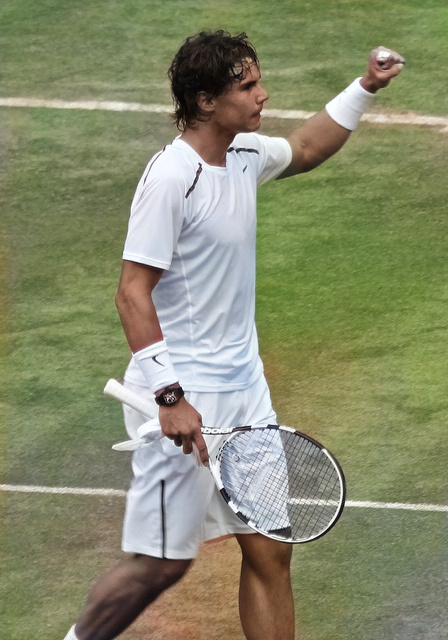}
&
    \includegraphics[width=0.2\textwidth]{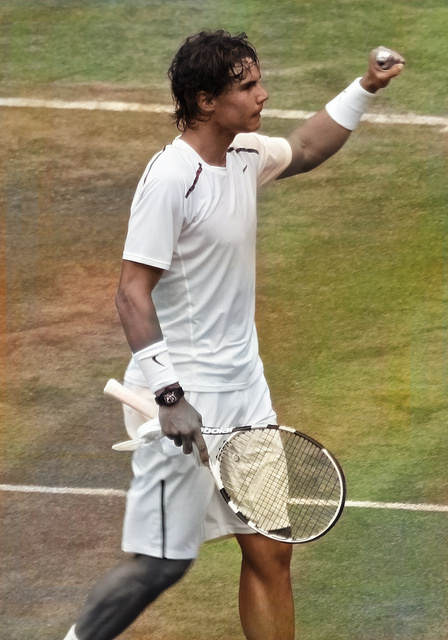}
&
    \includegraphics[width=0.2\textwidth]{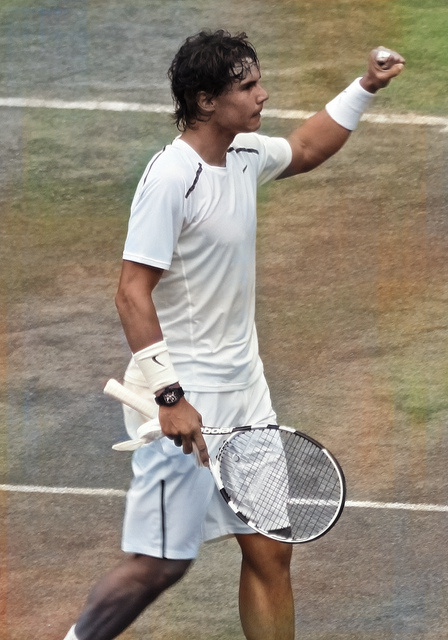}
&
    \includegraphics[width=0.2\textwidth]{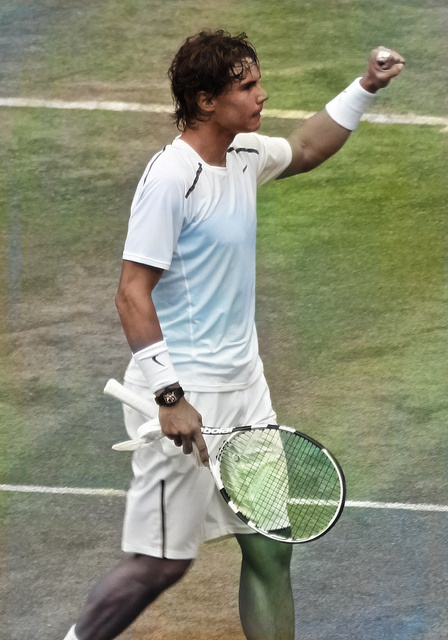}
\\
    \includegraphics[width=0.2\textwidth]{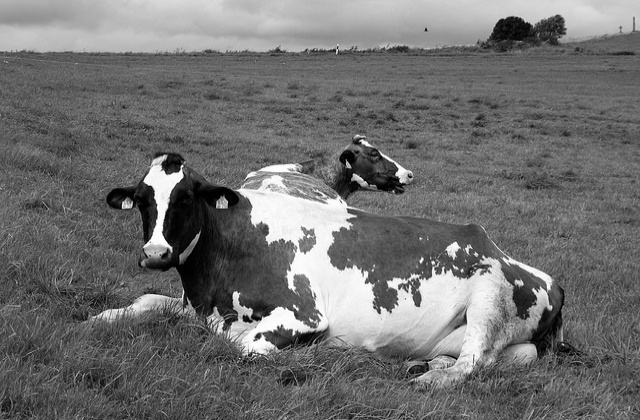}
&
    \includegraphics[width=0.2\textwidth]{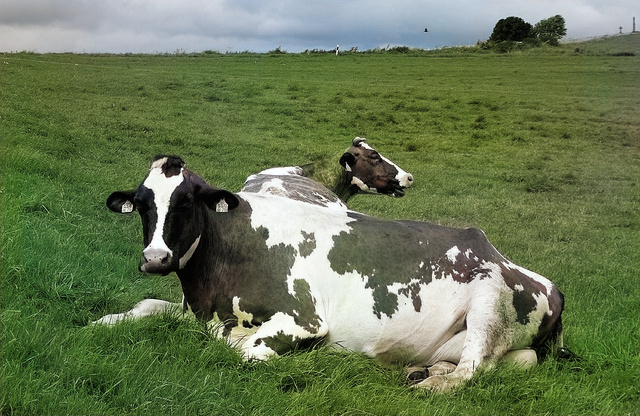}
&
    \includegraphics[width=0.2\textwidth]{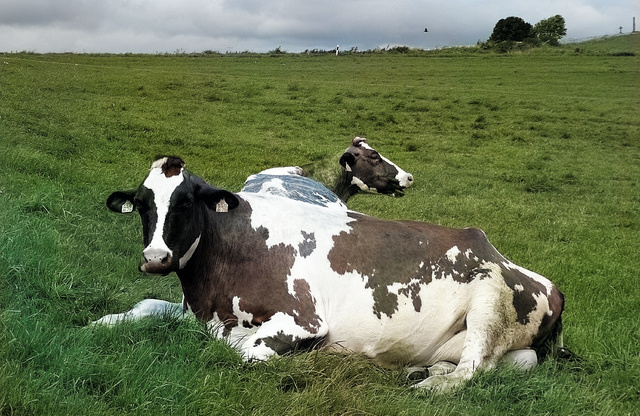}
&
    \includegraphics[width=0.2\textwidth]{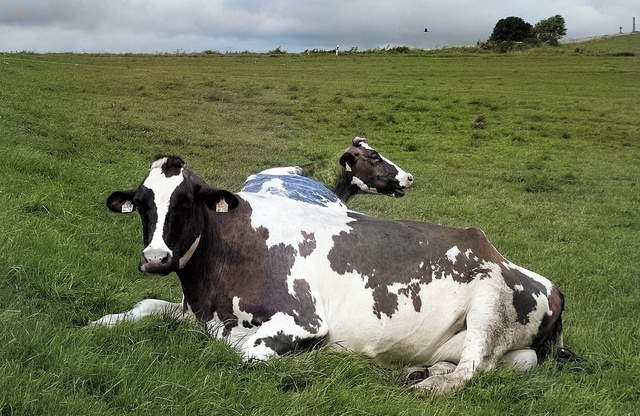}
&
    \includegraphics[width=0.2\textwidth]{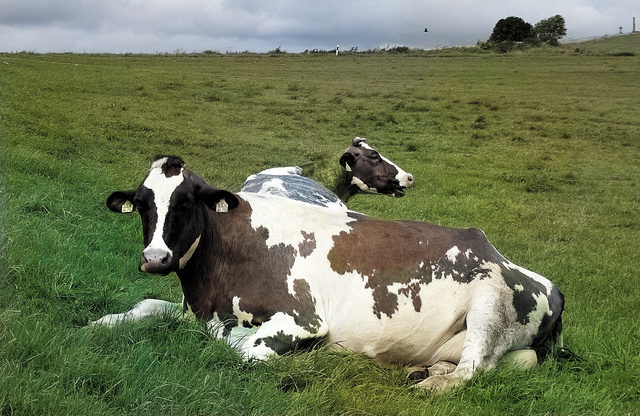}
\\
 Input   &
 RGB-L2  &
 YUV-L2  &
 Lab-L2  &
LabRGB-L2
\\
 \end{tabular}
 \begin{tabular}{ccc}
    \includegraphics[width=0.2\textwidth]{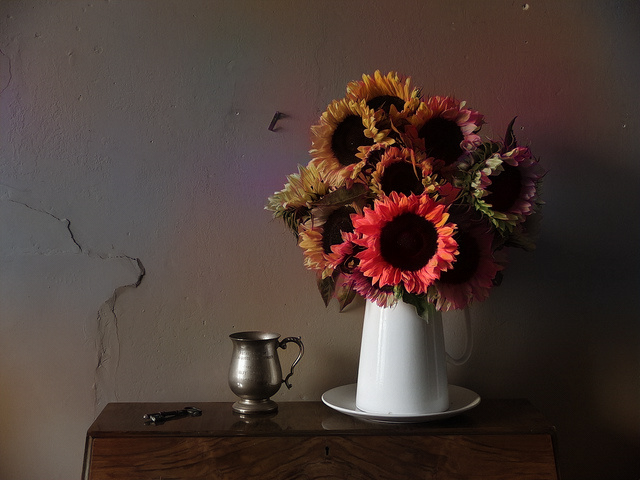}
&
    \includegraphics[width=0.2\textwidth]{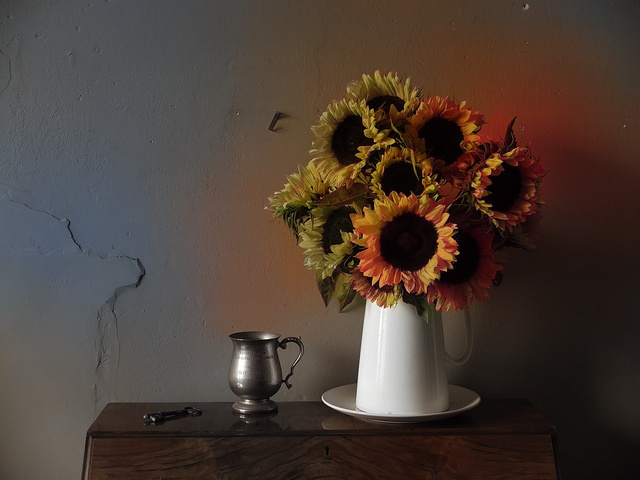}
&
    \includegraphics[width=0.2\textwidth]{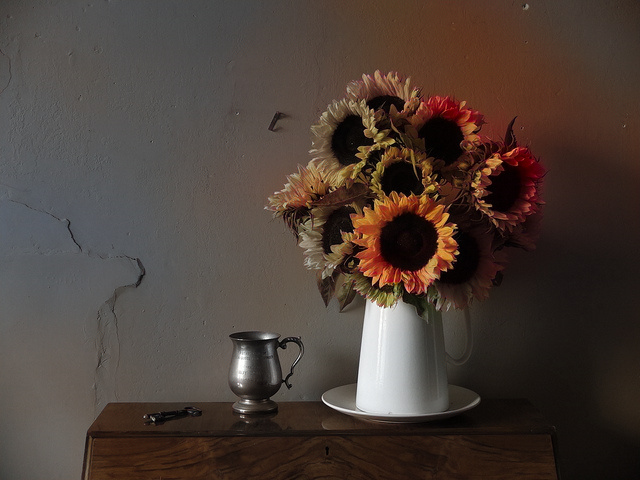}
\\
    \includegraphics[width=0.2\textwidth]{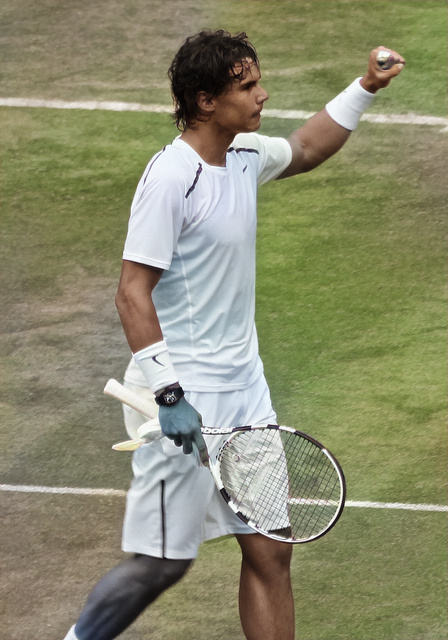}
&
    \includegraphics[width=0.2\textwidth]{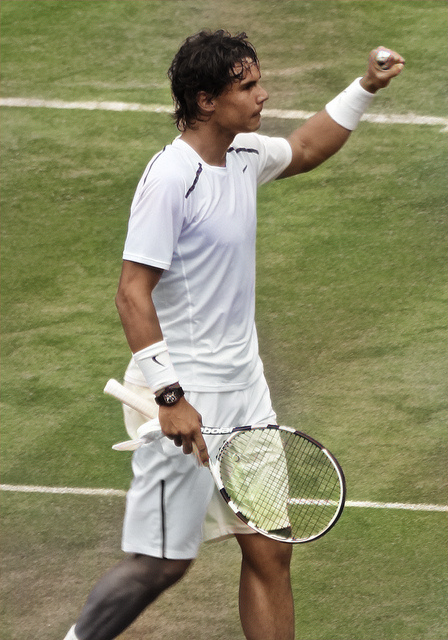}
&
    \includegraphics[width=0.2\textwidth]{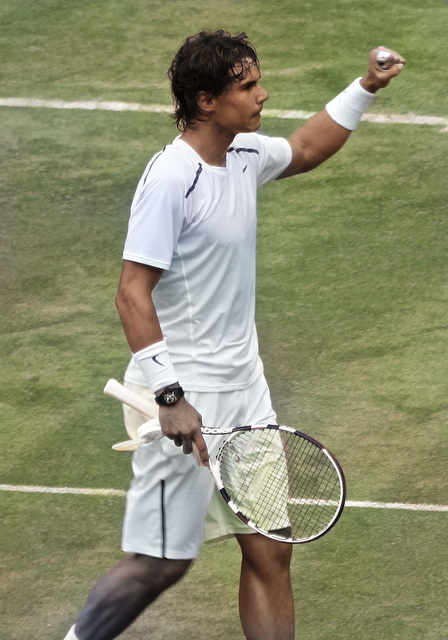}
\\
    \includegraphics[width=0.2\textwidth]{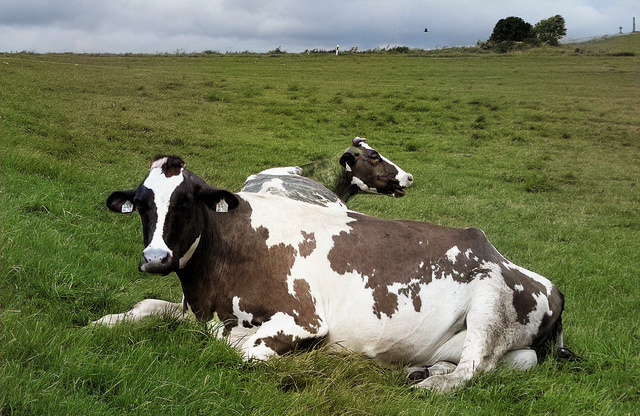}
&
    \includegraphics[width=0.2\textwidth]{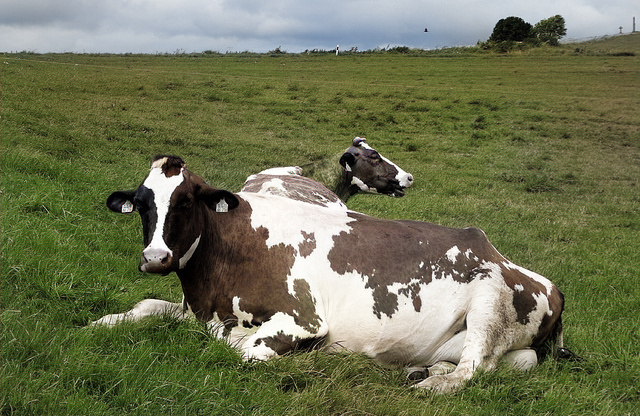}
&
    \includegraphics[width=0.2\textwidth]{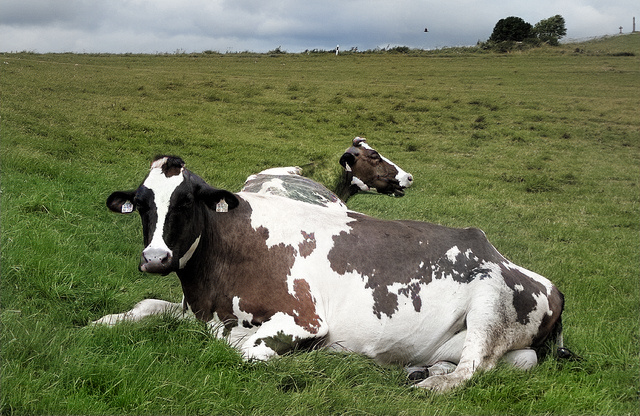}
\\
 RGB-LPIPS  &
 YUV-LPIPS  &
Lab-LPIPS
\\
 \end{tabular}
    \caption{Colorization results with different color spaces on images that contain small contours which lead to color bleeding. The two first rows are with L2 loss and the two last ones with VGG-based LPIPS. }\label{bleedingL2}
 \end{center} \end{figure}

One major problems in automatic colorization results come from color bleedings that occur as soon as contours are not strong enough.
Figure~\ref{bleedingL2} illustrates this problem in different contexts.
On the first row, the color from the flowers bleeds to the wall.
On the second row, the green of the grass bleeds to the shorts.
Finally, on the last row, the green of the grass bleeds to the neck of the background cow.
These effects are independent from the color space or the loss.
Some methods reduce this effect by introducing semantic information (e.g.,~\cite{vitoria2020chromagan}) or spatial localization (e.g.,~\cite{su2020instance}), while others achieve to reduce it by considering segmentation as an additional task (e.g.,~\cite{kong2021adversarial}).
Note that with the VGG-based LPIPS, Lab color space provides more realistic result on the tennisman image.

\begin{figure}[t]
\begin{center}
\begin{tabular}{ccccc}
    \includegraphics[width=0.2\textwidth]{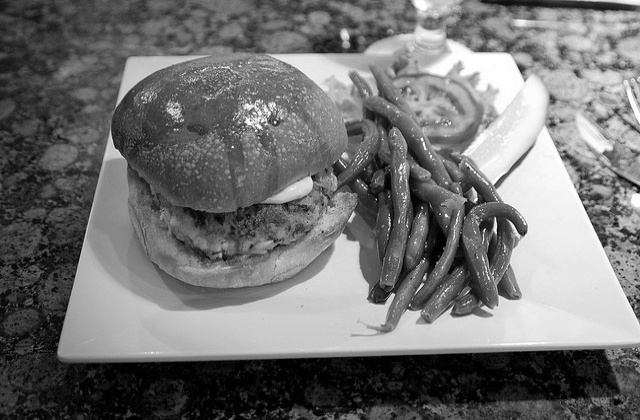}
& 
    \includegraphics[width=0.2\textwidth]{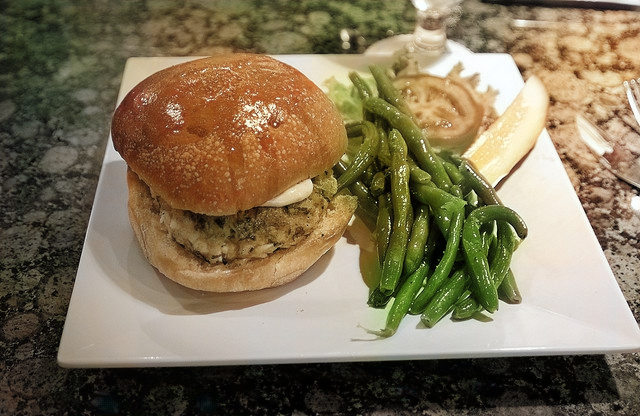}
& 
    \includegraphics[width=0.2\textwidth]{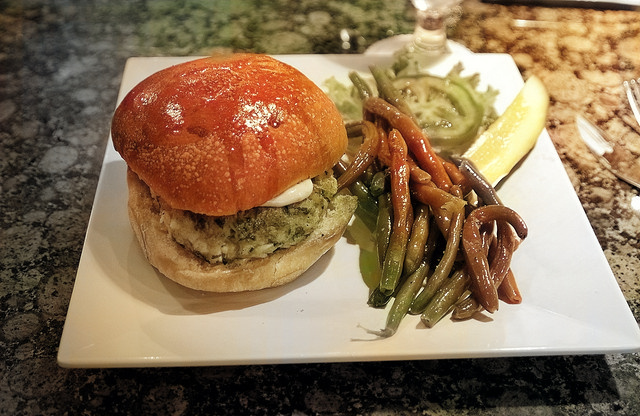}
& 
    \includegraphics[width=0.2\textwidth]{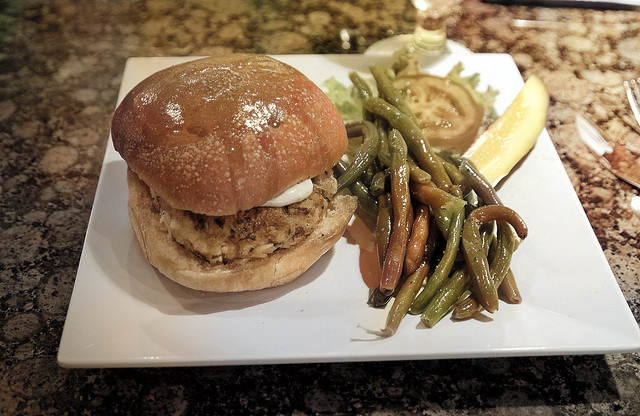}
& 
    \includegraphics[width=0.2\textwidth]{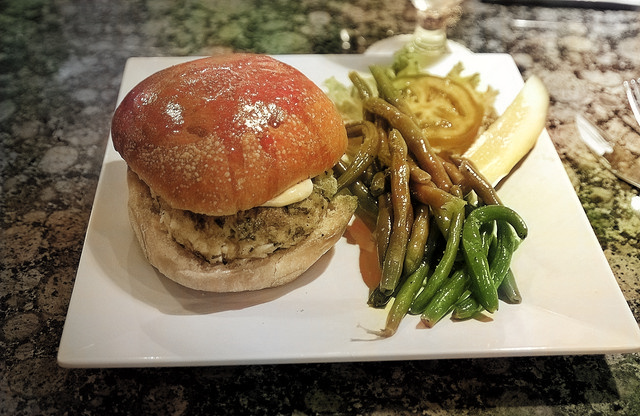}
\\
    \includegraphics[width=0.2\textwidth]{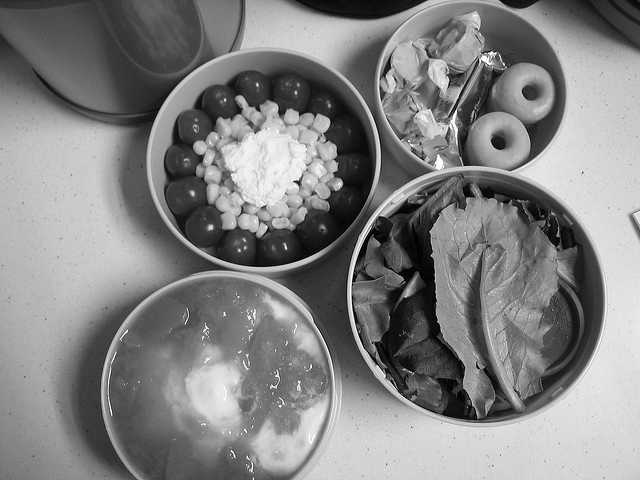}
&
    \includegraphics[width=0.2\textwidth]{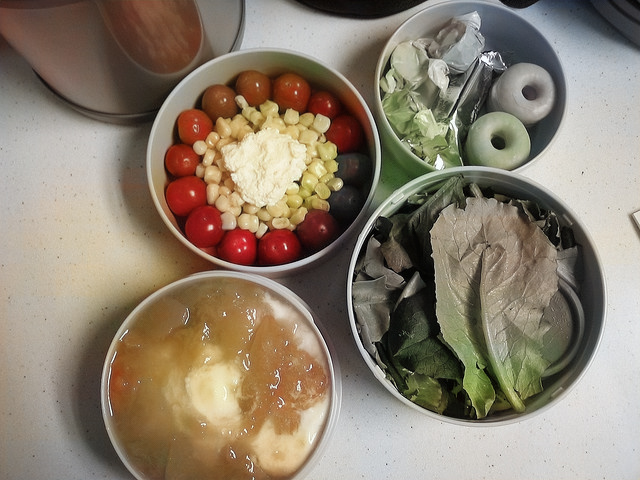}
& 
    \includegraphics[width=0.2\textwidth]{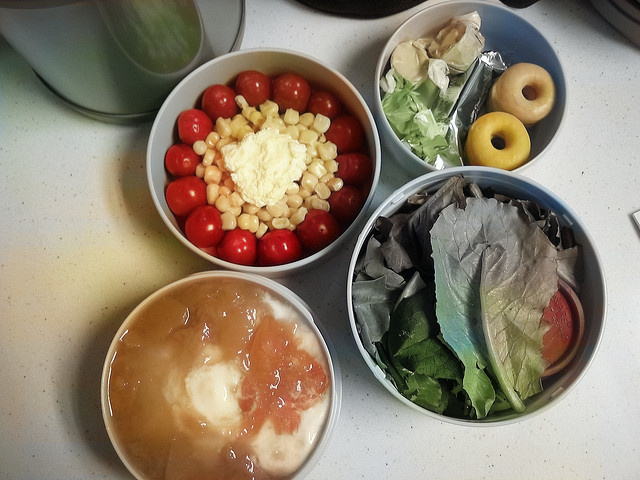}
& 
    \includegraphics[width=0.2\textwidth]{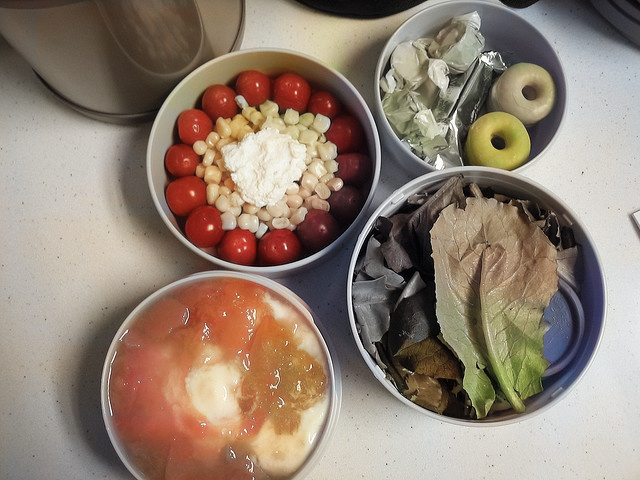}
& 
    \includegraphics[width=0.2\textwidth]{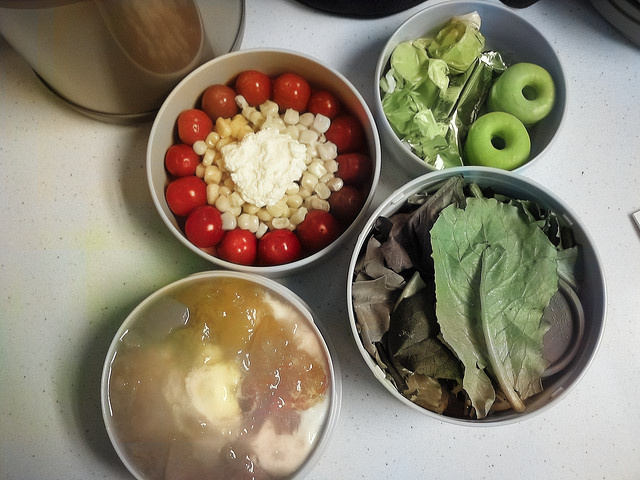}
\\
    \includegraphics[width=0.2\textwidth]{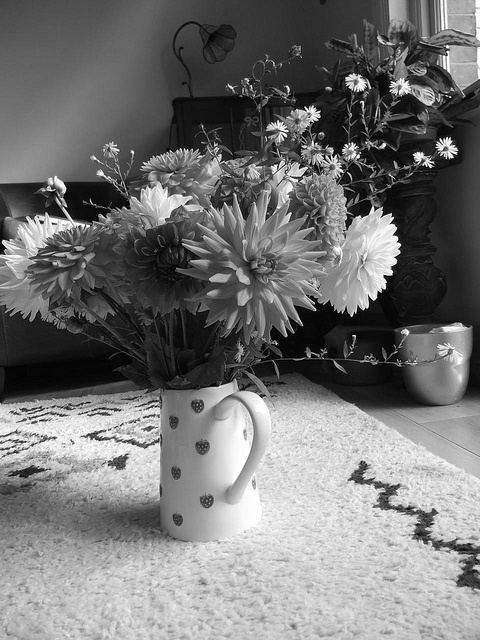}
& 
    \includegraphics[width=0.2\textwidth]{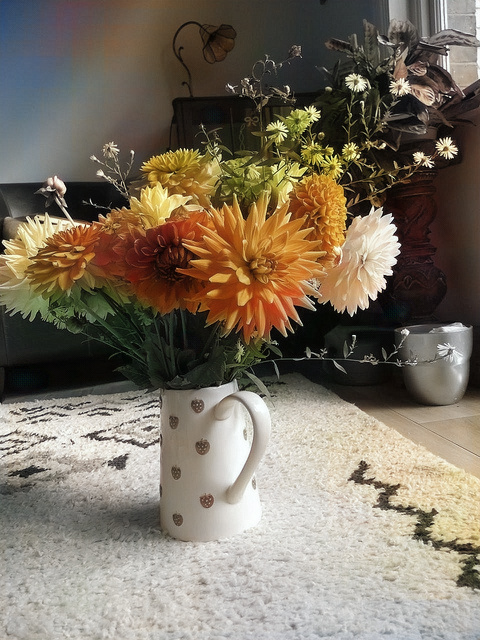}
& 
    \includegraphics[width=0.2\textwidth]{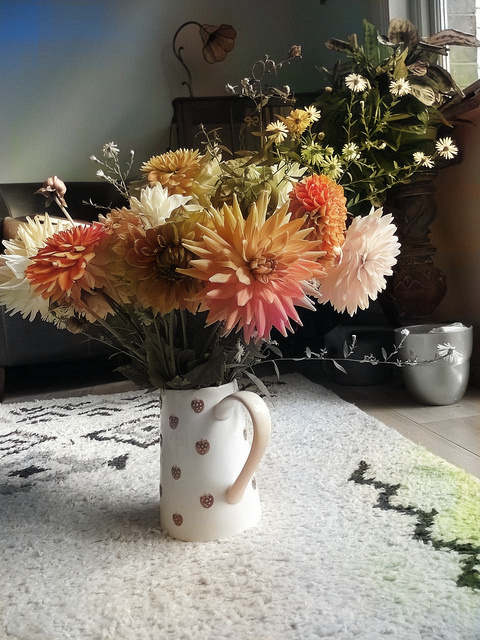}
& 
    \includegraphics[width=0.2\textwidth]{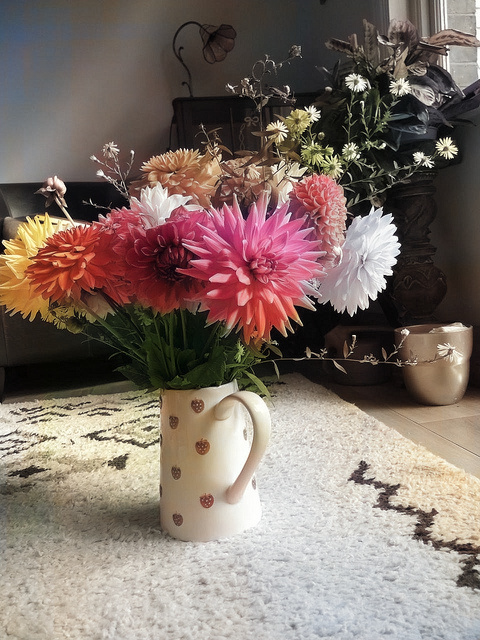}
& 
    \includegraphics[width=0.2\textwidth]{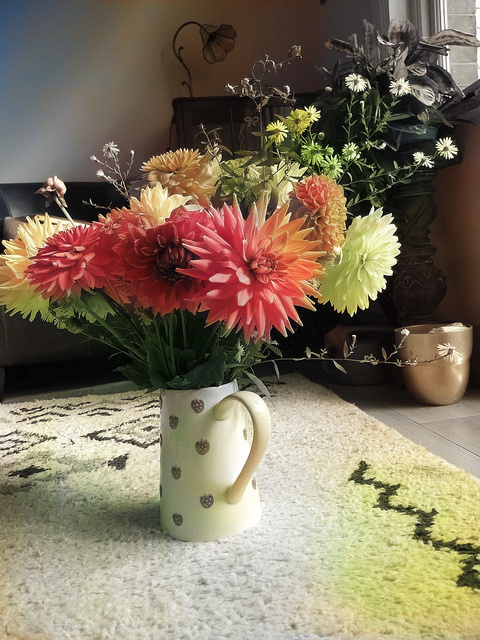}
\\
 Input  &
 RGB-L2  & 
 YUV-L2  & 
 Lab-L2  & 
LabRGB-L2
\\
 \end{tabular}
 \begin{tabular}{ccc}
    \includegraphics[width=0.2\textwidth]{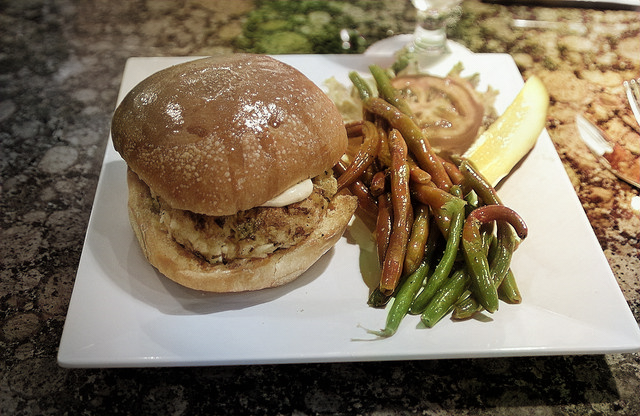}
& 
    \includegraphics[width=0.2\textwidth]{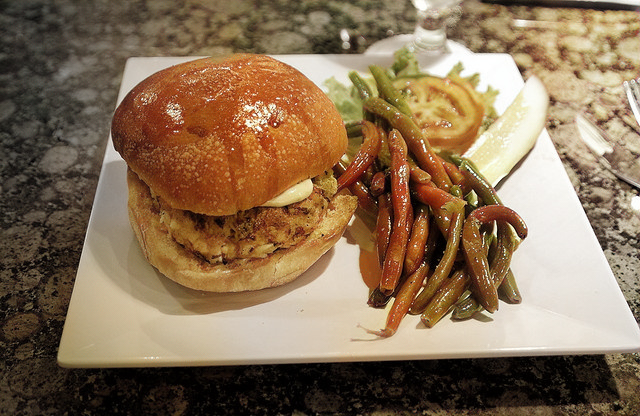}
& 
    \includegraphics[width=0.2\textwidth]{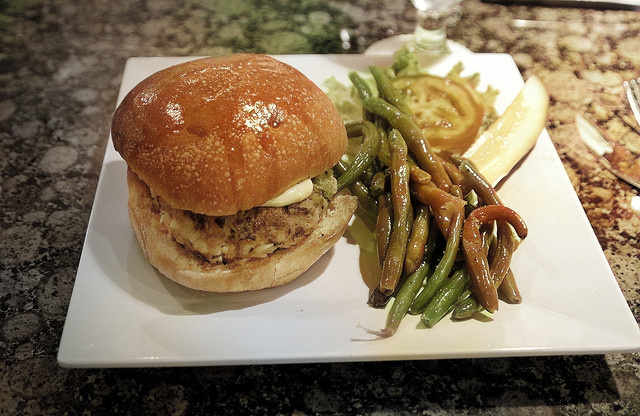}
\\
    \includegraphics[width=0.2\textwidth]{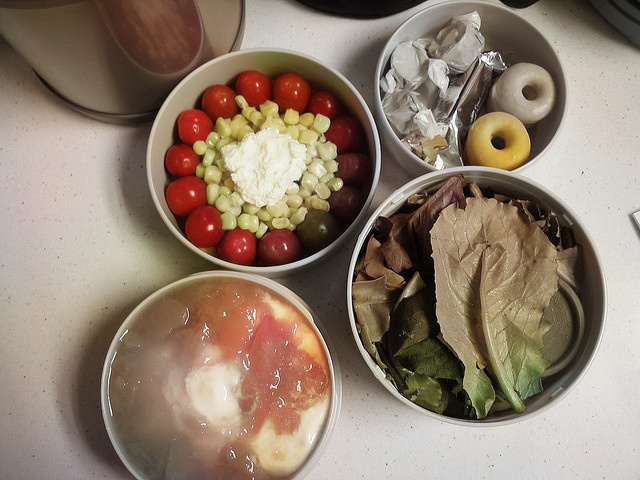}
& 
\includegraphics[width=0.2\textwidth]{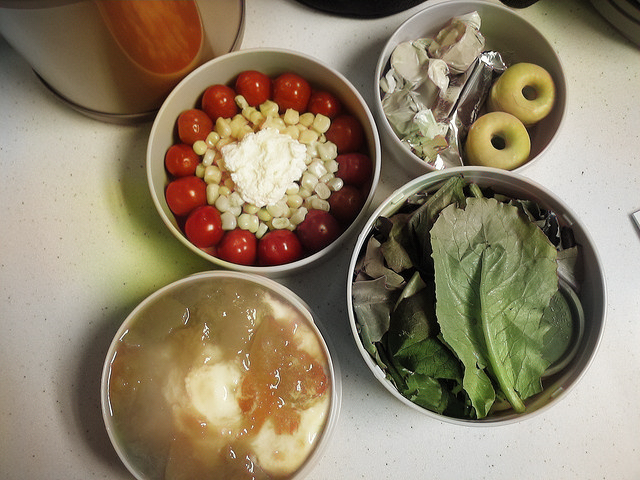}
& 
\includegraphics[width=0.2\textwidth]{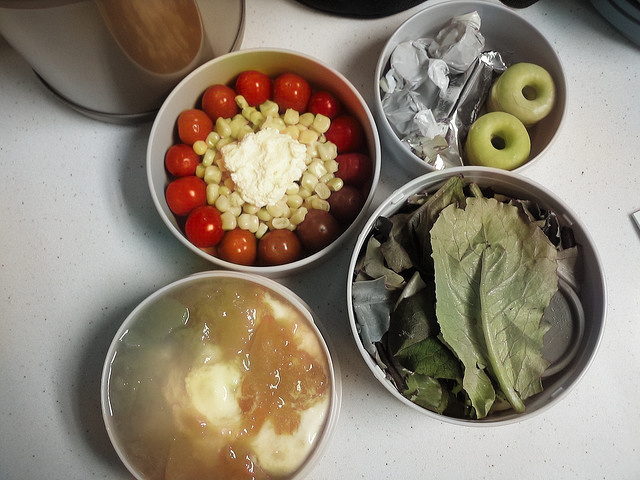}
\\
    \includegraphics[width=0.2\textwidth]{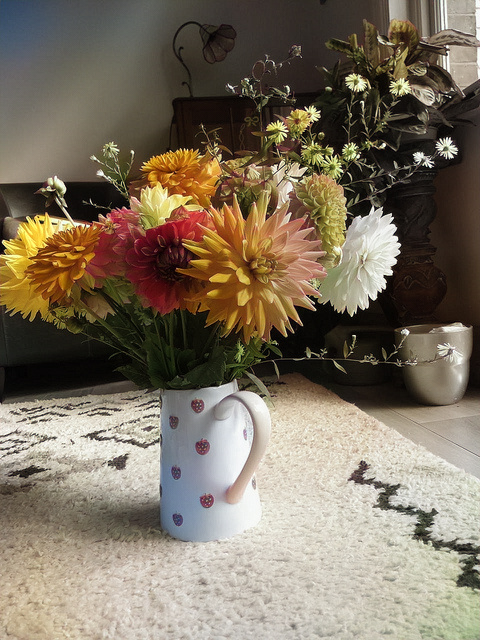}
& 
    \includegraphics[width=0.2\textwidth]{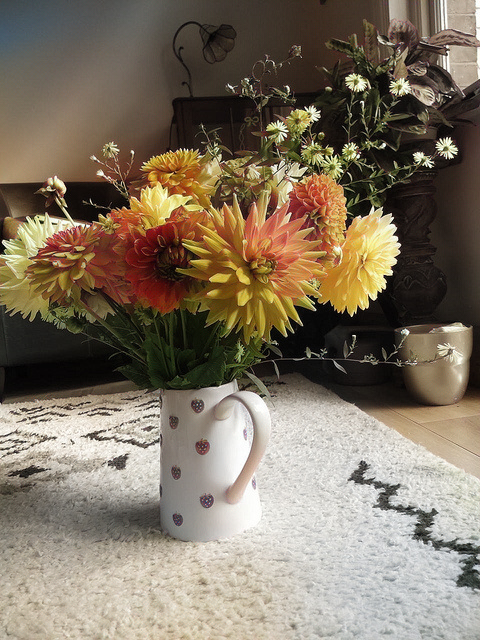}
& 
    \includegraphics[width=0.2\textwidth]{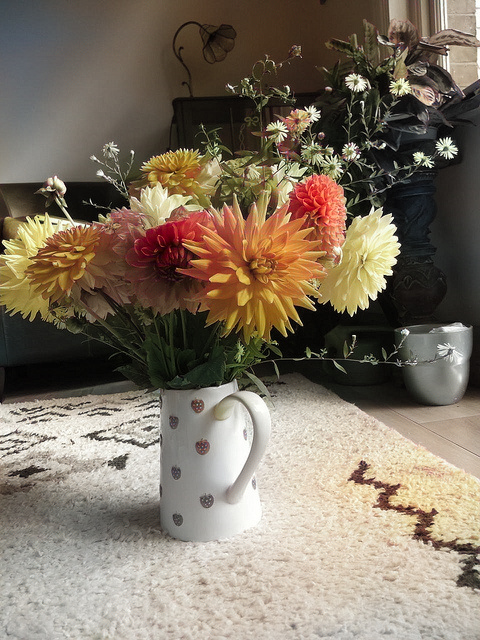}
\\
 RGB-LPIPS  & 
 YUV-LPIPS  & 
Lab-LPIPS
\\
 \end{tabular}
 \caption{Colorization results with different color spaces on images that contain several small objects which end up with different colors depending on the color spaces used. The three first rows are with L2 loss and the three last ones with VGG-based LPIPS.}\label{manyobjectsL2}
 \end{center} \end{figure}

Finally, Figure~\ref{manyobjectsL2} presents colorization of images containing many different objects.
We see that final colors might be dependent on the color spaces, and are more diverse and colorful with Lab color space.
LabRGB strategy with L2 loss is probably the more realistic, statement that holds with the VGG-based LPIPS.

The qualitative evaluation does not point to the same conclusion as the quantitative one.
According to Table~\ref{table:Quantitativeresults}, the best colorization is obtained for YUV color space.
However, the qualitative analysis shows that even if in some cases colors are brighter and more saturated in other ones it creates unpredictable color stains (yellowish and blueish).
This raises the question on the necessity to design specific metrics for the colorization task, which should be combined with user studies.
Also,in the qualitative evaluation one can observe that when working with LabRGB instead of Lab the overall colorization result looks more stable and homogeneous as opposed to what is concluded in the quantitative evaluation.

\noindent \underline{Summary of qualitative analysis}: Our analysis leads us to the following conclusions:
\begin{itemize}
\item There is no major difference in the results regarding the color space that is used.
\item YUV color space sometimes generates color artifacts that are hardly predictable. This is probably due to clipping that is necessary to remain in the color space range of values.
\item More realistic and consistent results are obtained when losses are computed in the RGB color space.
\item There is no evidence justifying why most colorization methods in the literature choose to work with Lab. One can assume that this is mainly done to ease the colorization problem by working in a perceptual luminance-chrominance color space. In addition, differentiable color conversion libraries were not available up to 2020 to apply a strategy as in Figure~\ref{fig:strategy}(c). In fact, the qualitative results show that when training on RGB the luminance reconstruction is satisfying in all examples. Hence, there is no obvious reason why not to work directly in RGB color space.
\item Same conclusions hold with different losses.
\end{itemize}

\section{Generalization to Archive Images}
\label{sec:archive}

\begin{figure}[t]
\begin{center}
\begin{tabular}{cccccccc}
    \includegraphics[width=0.135\textwidth]{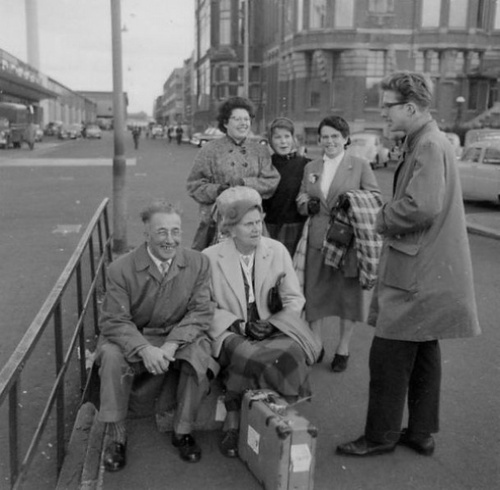}
& 
    \includegraphics[width=0.135\textwidth]{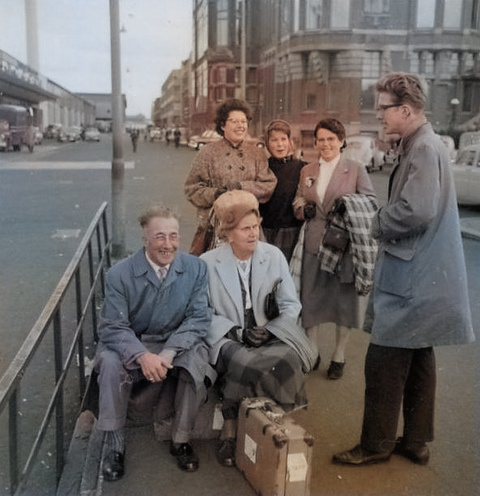}
& 
    \includegraphics[width=0.135\textwidth]{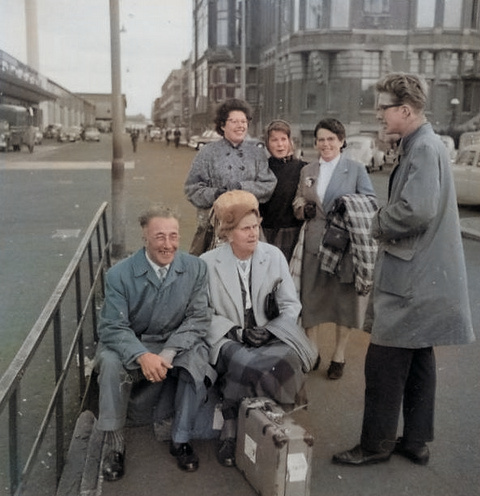}
& 
    \includegraphics[width=0.135\textwidth]{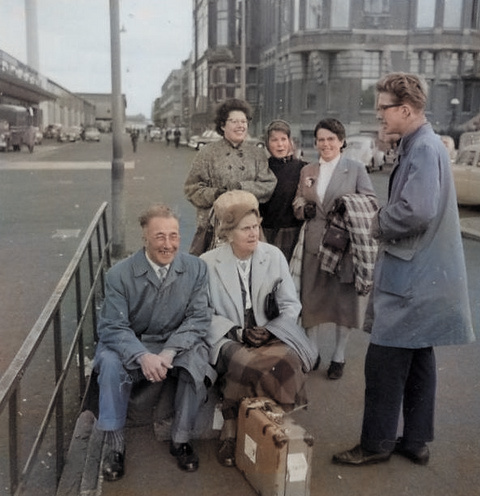}
& 
    \includegraphics[width=0.135\textwidth]{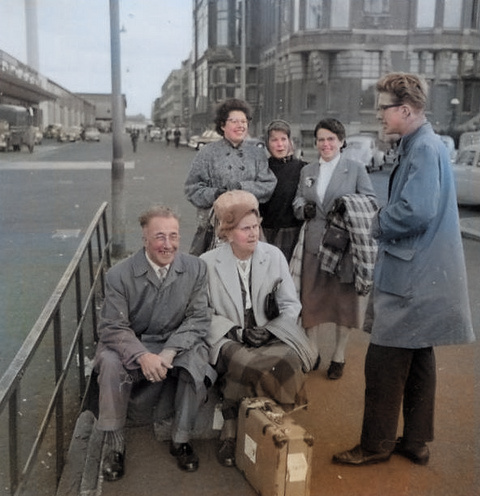}
& 
    \includegraphics[width=0.135\textwidth]{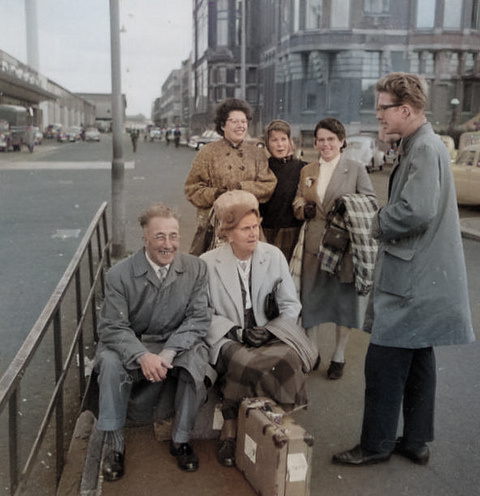}
& 
    \includegraphics[width=0.135\textwidth]{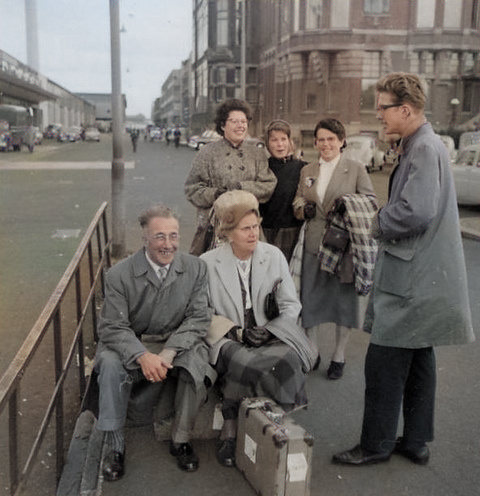}
& 
    \includegraphics[width=0.135\textwidth]{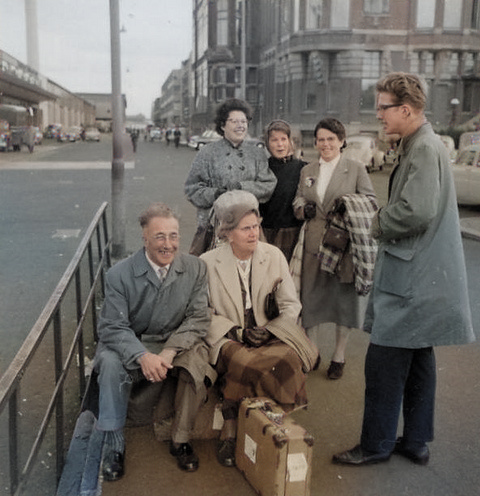}
\\

    \includegraphics[width=0.135\textwidth]{./img/gray_000000034843.jpg}
& 
    \includegraphics[width=0.135\textwidth]{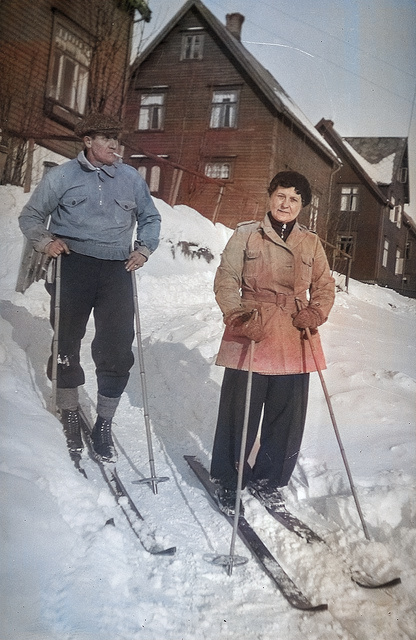}
& 
    \includegraphics[width=0.135\textwidth]{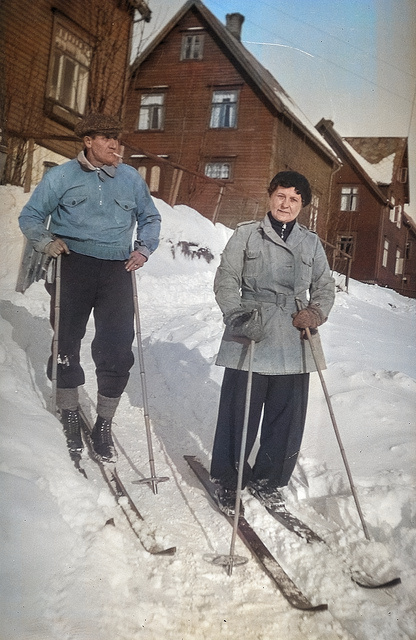}
& 
    \includegraphics[width=0.135\textwidth]{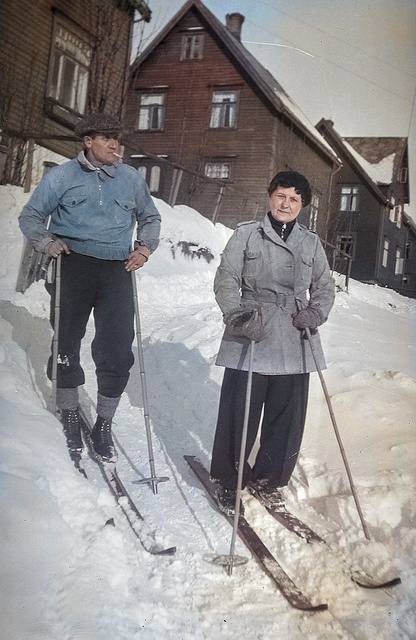}
& 
    \includegraphics[width=0.135\textwidth]{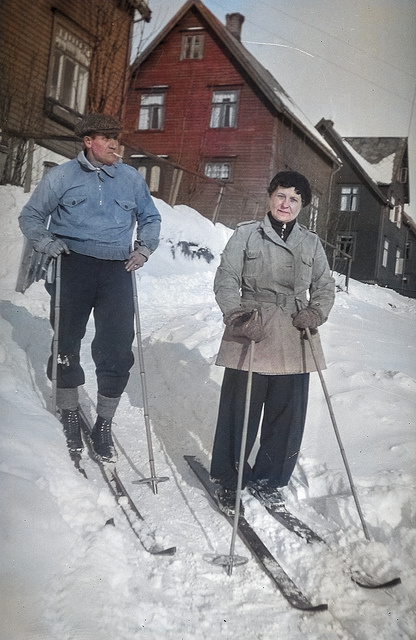}
& 
    \includegraphics[width=0.135\textwidth]{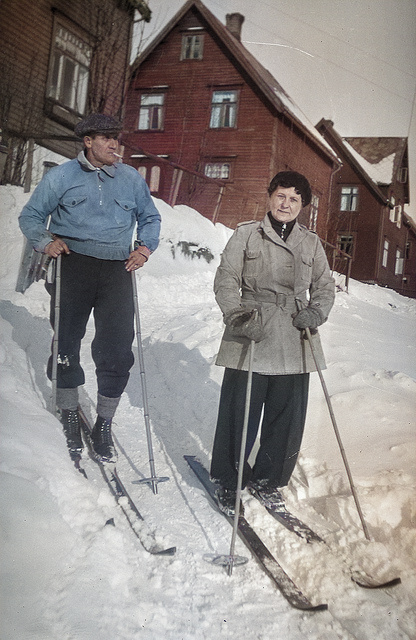}
& 
    \includegraphics[width=0.135\textwidth]{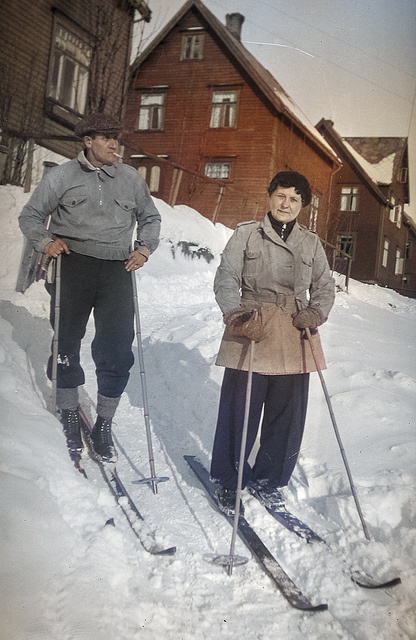}
& 
    \includegraphics[width=0.135\textwidth]{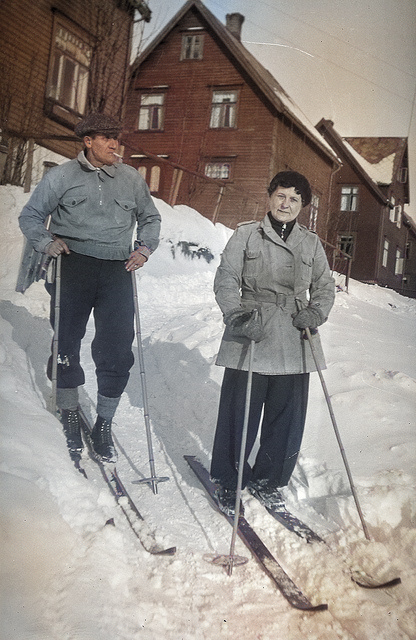}
\\
    \includegraphics[width=0.135\textwidth]{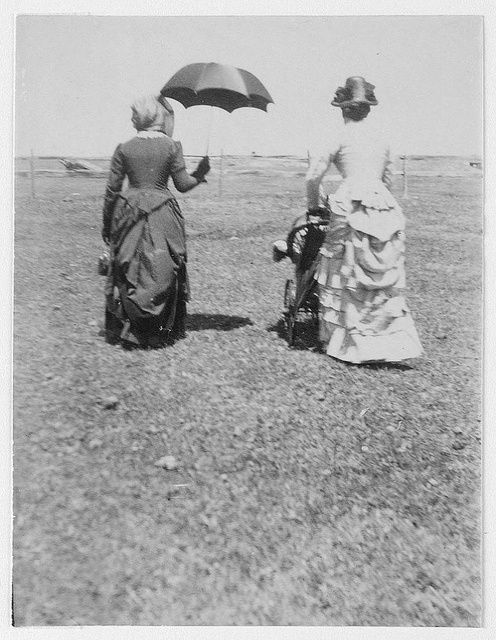}
&
    \includegraphics[width=0.135\textwidth]{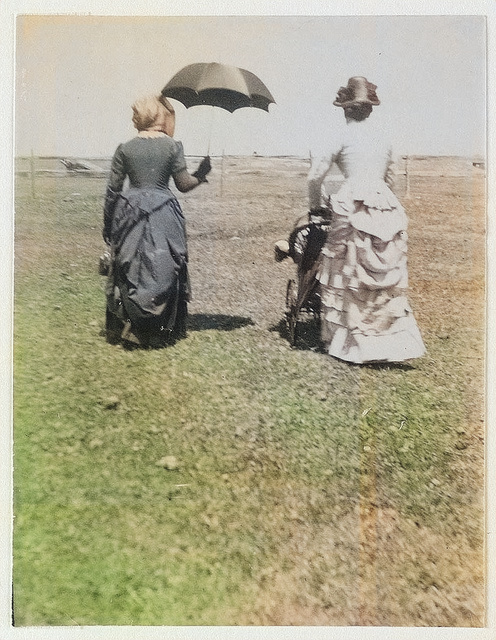}
& 
    \includegraphics[width=0.135\textwidth]{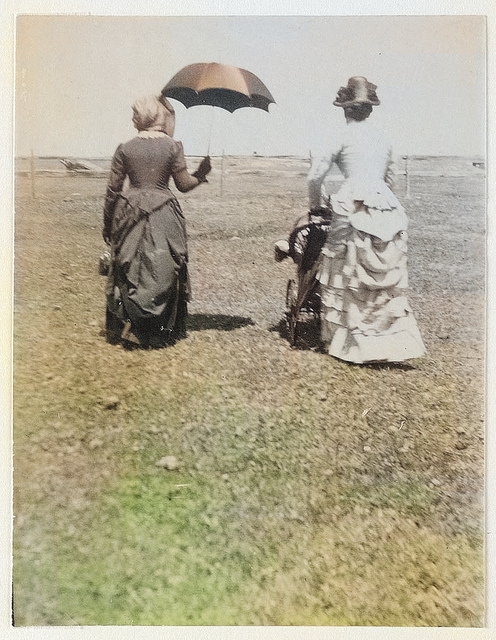}
& 
    \includegraphics[width=0.135\textwidth]{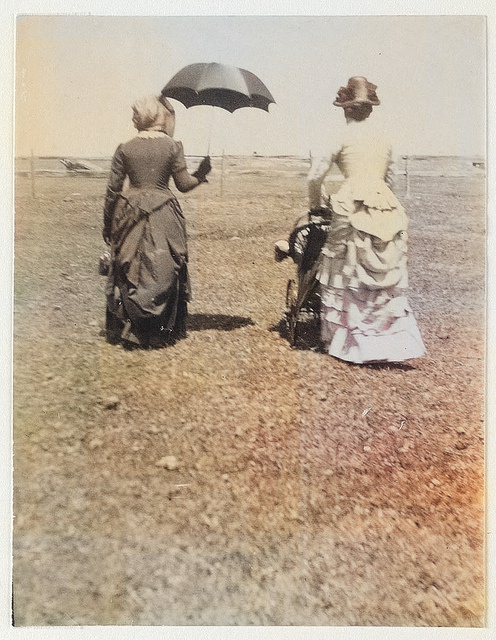}
& 
    \includegraphics[width=0.135\textwidth]{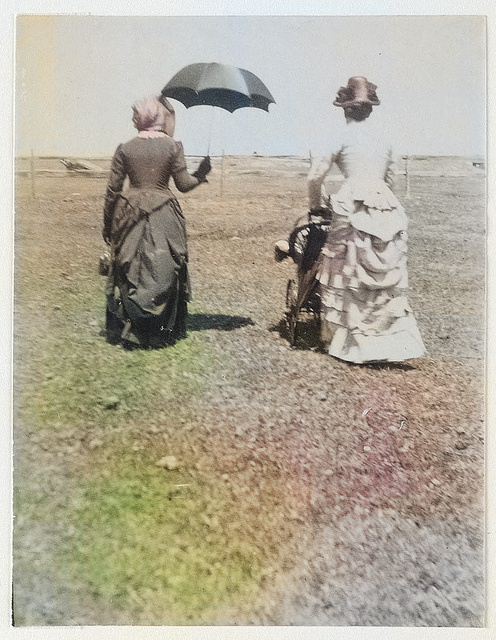}
& 
    \includegraphics[width=0.135\textwidth]{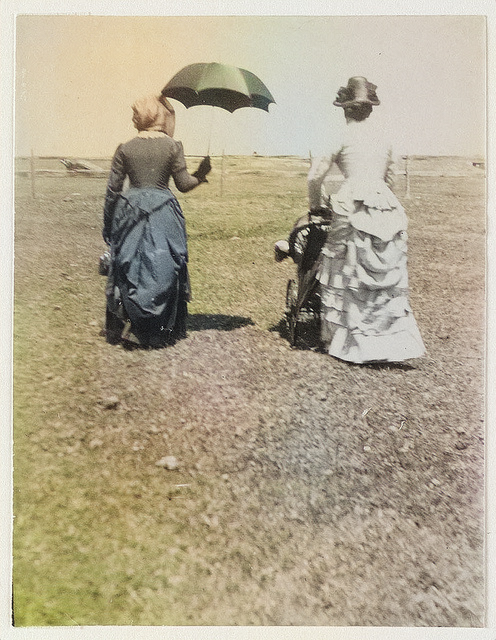}
& 
    \includegraphics[width=0.135\textwidth]{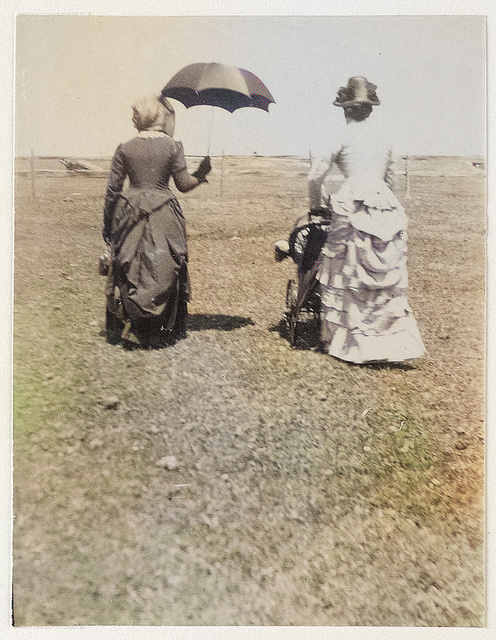}
& 
    \includegraphics[width=0.135\textwidth]{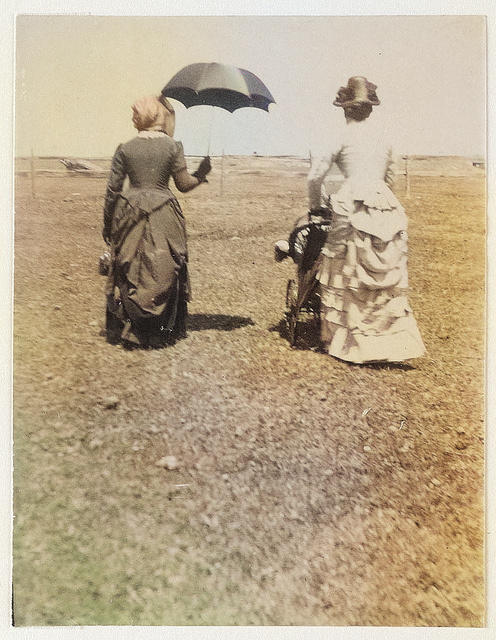}
\\
    \includegraphics[width=0.135\textwidth]{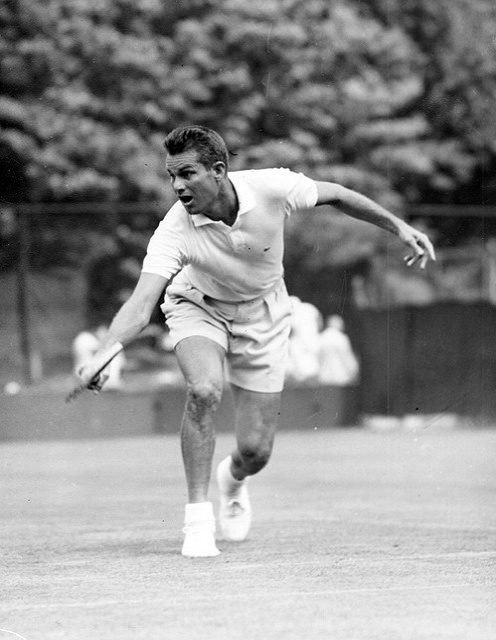}
& 
    \includegraphics[width=0.135\textwidth]{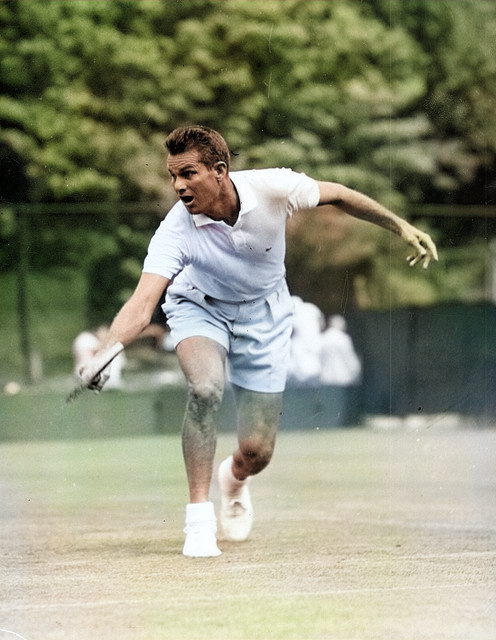}
& 
    \includegraphics[width=0.135\textwidth]{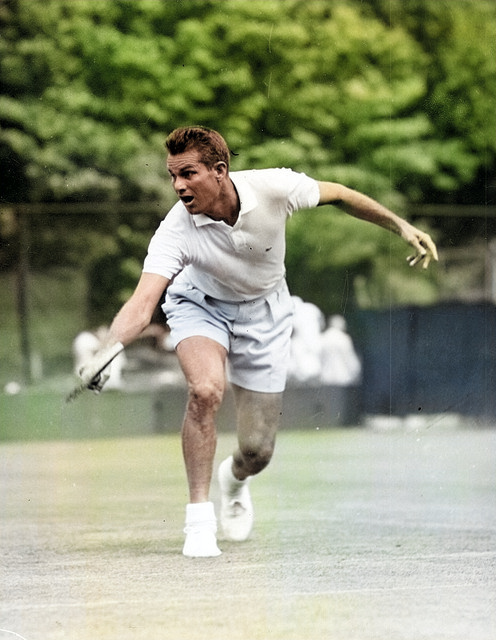}
& 
    \includegraphics[width=0.135\textwidth]{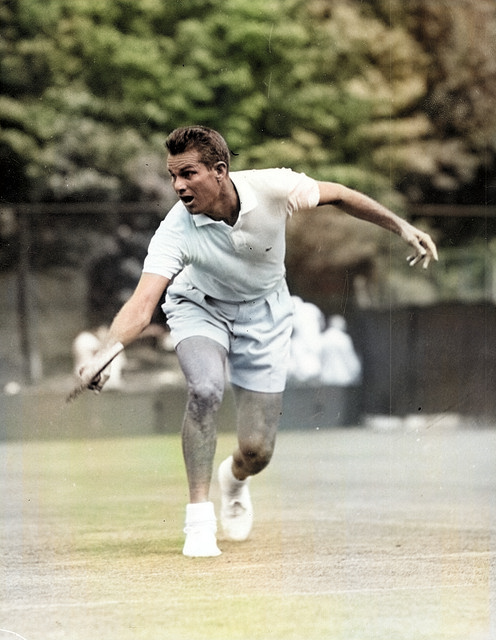}
& 
    \includegraphics[width=0.135\textwidth]{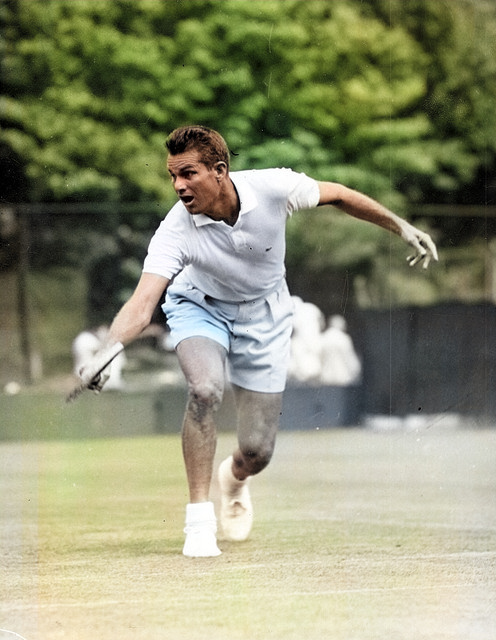}
& 
    \includegraphics[width=0.135\textwidth]{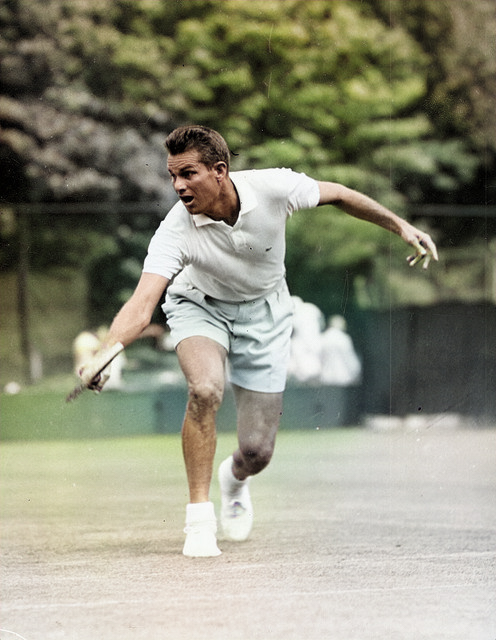}
& 
    \includegraphics[width=0.135\textwidth]{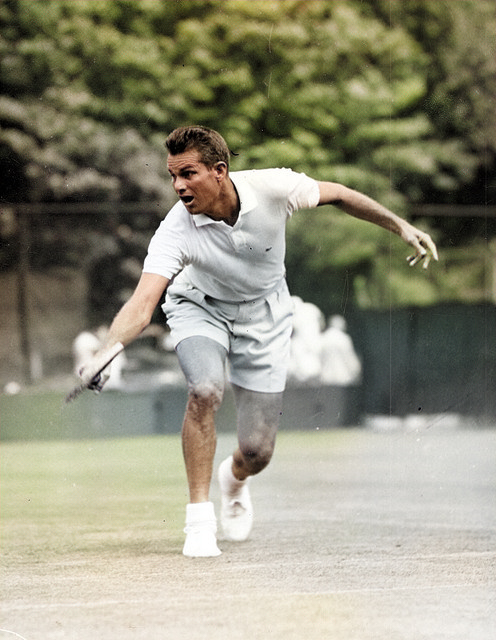}
& 
    \includegraphics[width=0.135\textwidth]{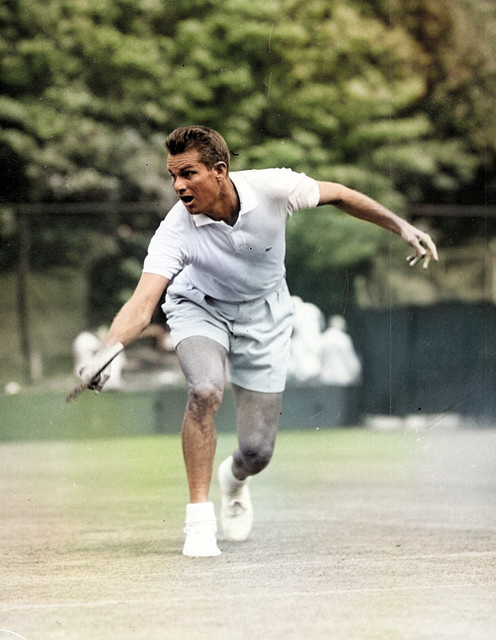}
\\
    \includegraphics[width=0.135\textwidth]{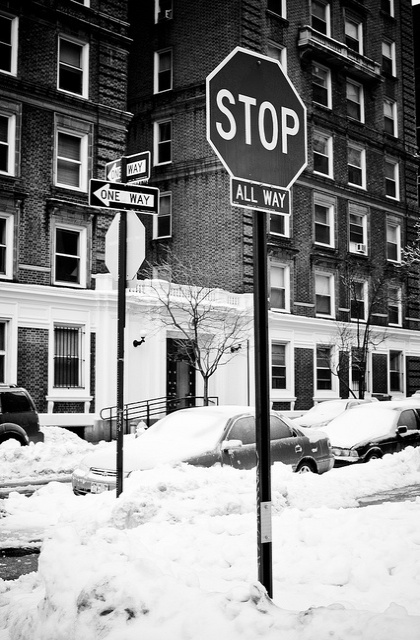}
& 
    \includegraphics[width=0.135\textwidth]{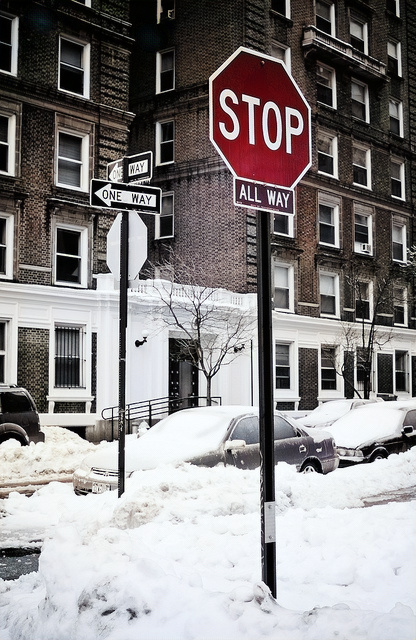}
& 
    \includegraphics[width=0.135\textwidth]{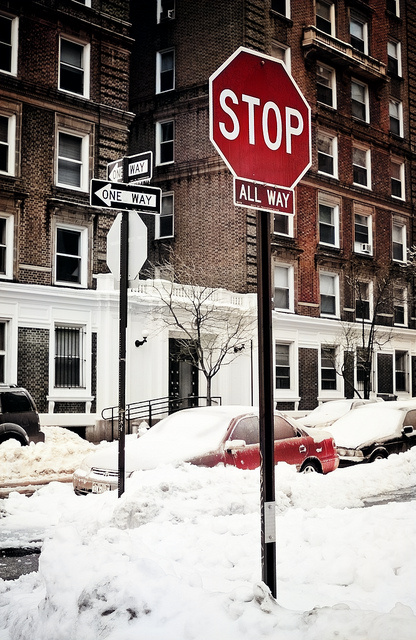}
& 
    \includegraphics[width=0.135\textwidth]{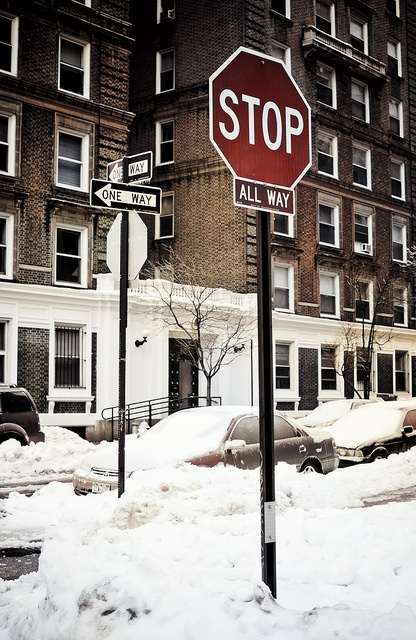}
& 
    \includegraphics[width=0.135\textwidth]{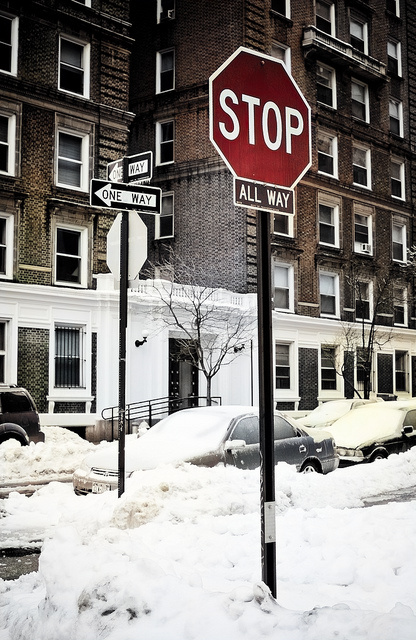}
& 
    \includegraphics[width=0.135\textwidth]{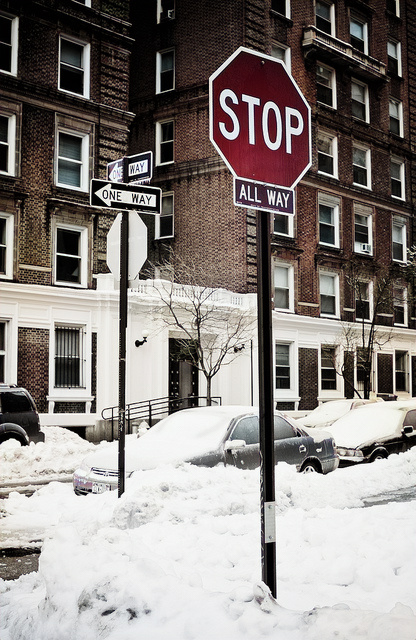}
& 
    \includegraphics[width=0.135\textwidth]{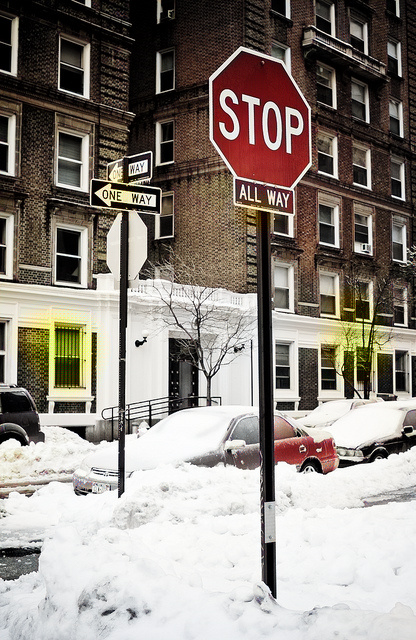}
& 
    \includegraphics[width=0.135\textwidth]{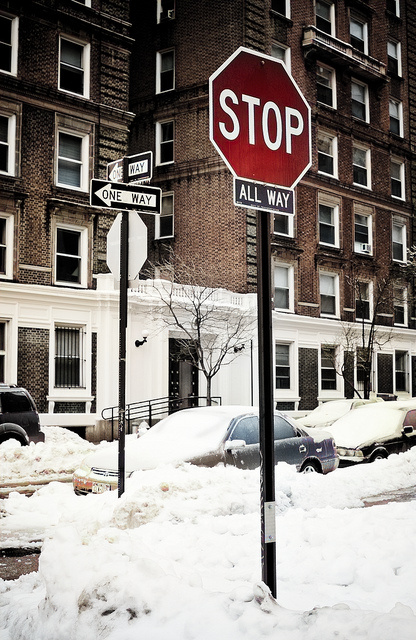}
\\
Input & RGB & YUV & Lab & LabRGB  & RGB & YUV  & Lab \\
& L2 & L2 & L2 & L2 & LPIPS & LPIPS & LPIPS
\\
\end{tabular}
\caption{Colorization results with different color spaces and L2 or VGG-based LPIPS on archive Black and White Images.}
\label{oldL2}
\end{center}
\end{figure}

Archive images present many artifacts due to acquisition methods (analog or numeric with different material qualities and manufacturing processes) and preservation conditions.
They lead to images with different resolutions, film grains, scratches and holes, flickering, etc.
Tools available for professional colorization enable artists to reach high level quality images but require long human intervention.
Current pipeline for professional colorization usually starts with restoration: denoising, deblurring, completion, super-resolution with off-the-shelf tools (\textit{e.g.} Diamant) and manual correction.
Next, images are segmented into objects and manually colorized by specialists with color spectrum that must be historically and artistically correct.

Automatic colorization methods could at least help professionals in the last step.
Very few papers in the literature tackle old black and white images' colorization.
In deep learning-based approaches, \cite{vitoria2020chromagan, antic2019deoldify} present some results on Legacy Black and White Photographs while~\cite{luo2020time} restores and colorizes old black and white portraits.
\cite{wan2020bringing} focuses on the restoration of old photos by training two variational autoencoders (VAE) to project clean and old photos to two latent spaces and to learn the translation between these latent spaces on synthetic paired data.
Old photos are synthesized using Pascal VOC dataset's images.

{Figure~\ref{oldL2} presents some results obtained by applying the networks trained in this chapter on archive images. As we can observe on second, third and fourth rows, while on clean images sky and grass are often well colorized, it is not the case on archive images. This is probably due to the grain and noise in these images. Similarly the skin of persons is not as well colorized as in clean images. Color bleeding is here again a real issue. On the other hand, for objects with strong contours that were present in the database (e.g. stop sign), the colorization works very well. This indicates the importance on training or fine tuning on images that are related to the purpose of the network (many of the objects present in old black and white photos are not well represented with the most often used datasets).}

\section{Conclusion}

This chapter has presented the role of the color spaces on automatic colorization with deep learning.
Using a fixed standard network, we have shown, qualitatively and quantitatively, that the choice of the right color space is not straightforward and might depend on several factors such as the architecture or the type of images.
With our architecture, best quantitative results are obtained in YUV while qualitative results rather teach us to compute losses in the RGB color space.
We therefore argue that most efforts should be made on the architecture design.
Furthermore, for all methods the final step consists in clipping final values to fit in the RGB color cube.
This abrupt operation sometimes leads to artifacts with saturated pixels.
An interesting topic for future research would be to learn a model that learns a projection into the color cube while preserving good image quality, similar to the geometric model from~\cite{pierre:hal-01168528}.
Future works should also include the development of methods that would give the possibility to produce several outputs in the same trend as HistoGAN~\cite{afifi2021histogan}.
Finally, if the purpose of colorization is often to enhance old black and white images, research papers rarely focus on this application.
Strategies for better training or transfer learning must be developed in the future along with complete architectures that perform colorization together with other quality improvement methods such as super resolution, denoising or deblurring.

\section*{Acknowledgements}

This study has been carried out with financial support from the French Research Agency through the PostProdLEAP project (ANR-19-CE23-0027-01) and from the EU Horizon
2020 research and innovation programme NoMADS (Marie Skłodowska-Curie grant agreement No 777826). This chapter was written together with another chapter of the current handbook, called \textit{Analysis of Different Losses for Deep Learning Image Colorization}. All authors have contributed to both chapters.

\bibliographystyle{apalike}
\bibliography{chapter_color}

\end{document}